\pdfoutput=1
\documentclass[10pt,twocolumn,letterpaper]{article}

\usepackage[pagenumbers]{iccv} 

\usepackage[accsupp]{axessibility}
\definecolor{iccvblue}{rgb}{0.21,0.49,0.74}
\usepackage[pagebackref,breaklinks,colorlinks,allcolors=iccvblue]{hyperref}

\newcommand{\pipeline}{FreeFine}
\newcommand{\Attention}{Temporal Contextual Attention}
\newcommand{\attention}{TCA}

\usepackage{makecell}
\usepackage{multirow}
\usepackage{booktabs}
\usepackage{amsfonts}
\usepackage{colortbl}
\usepackage{stfloats}
\usepackage{xr}
\usepackage{xr-hyper}

\def\paperID{6045} 
\def\confName{ICCV}
\def\confYear{2025}

\title{Training-Free Geometric Image Editing on Diffusion Models}

\author{
Hanshen Zhu$^{1*}$\quad
Zhen Zhu$^{2*}$\quad
Kaile Zhang$^1$\quad
Yiming Gong$^2$\quad
Yuliang Liu$^1$ \quad
Xiang Bai$^{1\dagger}$ \\
$^1$Huazhong University of Science and Technology \\
$^2$University of Illinois at Urbana-Champaign \\
{\tt\small $^1$\{zhs\_china, klzhang, ylliu, xbai\}@hust.edu.cn} \\
{\tt\small $^2$\{zhenzhu4, yimingg8\}@illinois.edu}
}

\begin{document}

\twocolumn[{
\maketitle
\begin{center}
    \centering
    \captionsetup{type=figure}
    \includegraphics[width=\textwidth]{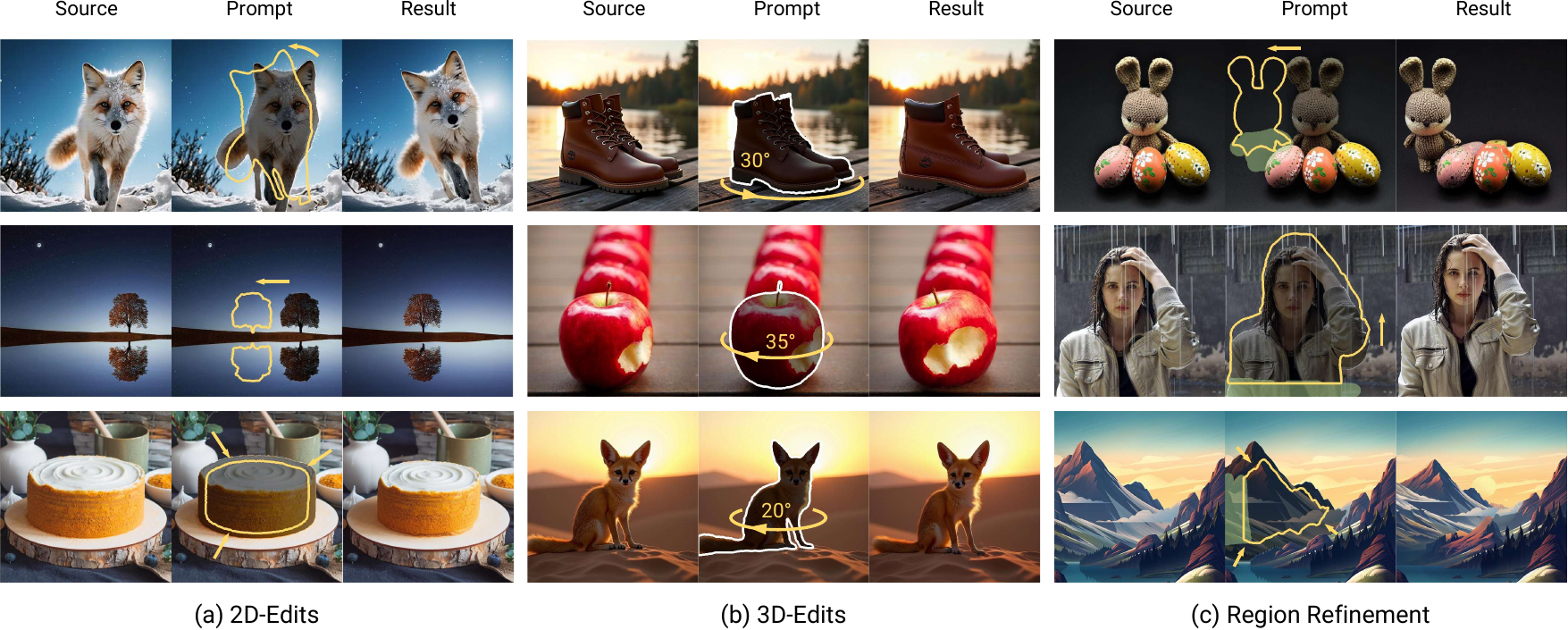}
    \captionof{figure}{Given an image and an editing instruction, our method precisely performs geometric edits while maintaining high fidelity and avoiding artifacts. Besides, our training-free framework achieves impressive structural completion and background generation. 
    }
    \vspace{10pt}
    \label{fig:teaser}
\end{center}
}]

\protect \renewcommand{\thefootnote}{\fnsymbol{footnote}}
\footnotetext[1]{Equal contribution. $^{\dag}$Corresponding author.}

\begin{abstract}
    We tackle the task of geometric image editing, where an object within an image is repositioned, reoriented, or reshaped while preserving overall scene coherence. Previous diffusion-based editing methods often attempt to handle all relevant subtasks in a single step, proving difficult when transformations become large or structurally complex. We address this by proposing a decoupled pipeline that separates object transformation, source region inpainting, and target region refinement. Both inpainting and refinement are implemented using a \textbf{training-free} diffusion approach, \textbf{FreeFine}. In experiments on our new \textbf{GeoBench} benchmark, which contains both 2D and 3D editing scenarios, FreeFine outperforms state-of-the-art alternatives in image fidelity,  and edit precision, especially under demanding transformations. 
    Code and benchmark are available at: \url{https://github.com/CIawevy/FreeFine}
    
\end{abstract}
    
\vspace{-15pt}

\section{Introduction}
\label{sec:intro}

Image generation models have made remarkable progress in producing photorealistic and detail-rich results~\cite{SD,SDXL,dalle3,imagen}. With these advancements, the community has shown growing interest in controllable image editing to enable users to manipulate existing images with both high fidelity and accuracy. In this paper, we address the task of repositioning, reorienting, or reshaping an object within an image (\eg, moving an object to a new location, rotating it in 3D, or changing its proportions) while preserving overall scene coherence, a task we refer to as {\bf geometric image editing}.


This problem requires solving multiple interdependent subtasks: (1) coarsely transforming the object to its desired location, (2) inpainting the source region to avoid artifacts, and (3) refining the relocated object to blend seamlessly with the background. Recent methods that support drag-based edits~\cite{Self-Guidance,MotionGuidance,RegionDrag,Dragon,dragdiffusion} typically address these goals with a single, unified objective, yielding compelling results for smaller or moderate transformations. However, they often struggle with large or geometrically complex transformations, possibly because balancing multiple subtasks in one optimization framework creates competing demands. For instance, strictly preserving the background may conflict with generating newly exposed object surfaces. Although it is challenging to pinpoint exactly how these objectives interact, the risk of artifacts in more extensive edits motivates us to take a \textbf{decoupled} strategy. Concretely, we separate geometric editing into three sequential steps, individually handling object transformation, source region inpainting, and target region refinement. This decomposition avoids the pitfalls of fusing significant structural changes and fine-grained touch-ups in a single loop, while allowing us to incorporate specialized off-the-shelf components, such as advanced depth estimators~\cite{DepthAnything} or video models~\cite{SV3D} for 3D transformations and dedicated inpainting models~\cite{Lama,MAT} for large-scale removals.

Within this pipeline, both inpainting and refinement can be cast as selectively altering regions based on spatial masks and generative priors. Instead of leveraging task-specific models~\cite{Lama,Brush,stableinpainting} that necessitate dedicated training, we adopt a \emph{training-free} diffusion-based strategy capable of handling both tasks with minimal tweaks, which combines three complementary modules: \emph{Temporally Contextual Attention} to balance self-attention with mask-guided attention over the course of diffusion steps, \emph{Local Perturbation} for selective noise injection that encourages substantial changes in user-defined areas, and \emph{Content-specified Generation} for text-driven local refinements. These modules collectively preserve global fidelity while generating plausible details in newly exposed or structurally incomplete regions.

To systematically evaluate geometric editing, we introduce \textbf{GeoBench}, a benchmark designed to test 2D and 3D transformation editing scenarios in varied degrees of difficulty. Each scenario includes a source image, one or more geometric editing instructions, and optional structural completion masks. We adopt popular generative metrics such as FID~\cite{FID}, as well as measures for subject/background consistency~\cite{vbench} and edit precision~\cite{geodiffuser}. 

\vspace{2pt}\noindent Our {\bf contributions} can be summarized as follows:
\begin{itemize}
    \item We present a decoupled geometric image editing pipeline that splits the editing process into object transformation, source region inpainting, and target region refinement, supporting both 2D and 3D transformations.
    \item We propose {\pipeline{}}, a \emph{training-free}, diffusion-based solution powered by \Attention{}, {Local Perturbation}, and {Content-specified Generation}. This design provides fine-grained control over the editing regions while maintaining global coherence.
    \item We introduce {GeoBench}, a benchmark specialized for evaluation on geometric image editing with diverse instructions and metrics. Our method significantly outperforms existing approaches in both large and small transformations.
\end{itemize}
\section{Related Work}
\noindent\textbf{Diffusion models.}~Diffusion models~\cite{DDPM,DDIM} synthesize images by iteratively denoising noisy inputs, with improvements such as non-Markovian sampling~\cite{DDIM} and latent diffusion~\cite{SD} greatly accelerating generation. Combined with large-scale language modeling, they power state-of-the-art text-to-image systems like DALL-E~\cite{dalle1,dalle2,dalle3} and Imagen~\cite{imagen}. Recent extensions include rectified flow models~\cite{rectifiedflow} for more efficient sampling. We refer readers to~\cite{diffusionsurvey} for broader coverage of research on diffusion models.

\vspace{1mm}\noindent\textbf{Geometric editing with generative models.}~Early methods for view manipulation and alignment~\cite{SpatialTransformer,DeepHomogrophy} lacked the ability to perform direct image editing. Advances in novel view synthesis and 3D representations~\cite{Nerf,GS,zero123,SV3D} enable more expressive geometric changes by reconstructing scenes in 3D space~\cite{ImageSculpting}. However, these methods typically require multi-view inputs or per-scene optimization, making them less practical for single-image, real-time editing. More recent work on single-image 3D reconstruction attempts to infer 3D shapes from sparse or single-view data~\cite{zero123,SV3D}, and can be seamlessly integrated into our framework: for larger 3D transformations, the object is temporarily “lifted” to 3D, manipulated, and reprojected back to 2D. Other approaches, such as pose transfer~\cite{ProgressivePoseTransfer,ProgressiveAlignPoseTransfer,PoseGuide,DefromGAN} or virtual try-on~\cite{VITON,CAP-VITON,DressInOrder}, focus on human-centric transformations and cannot generalize beyond that domain without significant modification.


Given source and target locations or constraints, a decent volume of research efforts focus on manipulating latents to minimize feature discrepancies.
They can be categorized by their guidance signals: (i) \emph{Point-based methods}~\cite{DragGAN,DragNoise,InstantDrag,dragdiffusion,dragapart(3D),stabledrag} accept user-defined points or anchors to indicate how parts of the object should move, (ii) \emph{Region-based methods}~\cite{geodiffuser,Self-Guidance,Diffusionhandles} operate on segmented areas to apply object-level edits, and (iii) \emph{Combined or more diverse signals}~\cite{Dragon,RegionDrag} fuse point and region information. Further variations exploit flow fields~\cite{MotionGuidance}, scene layouts~\cite{SceneDiffusion,readoutguidance,DesignEdit}, or higher-level editable elements~\cite{ImageElements}. 
Our method can optionally work with single-image 3D lifting~\cite{SV3D} to handle substantial viewpoint changes.
Meanwhile, our method remains compatible with diverse 2D editing operations without requiring multi-view data or specialized human-centric models.
\vspace{1mm}\noindent\textbf{Training-Free Inpainting and Restoration.}~
While some inpainting methods often rely on a specialized training procedure tailored to repair designated areas~\cite{Brush,stableinpainting}, recent works~\cite{GenerativeDiffusionPrior,FourierPrior,prior_restoration,attentiveEraser} have exploited the strong generative priors of large diffusion models for image inpainting and restoration without further training. For instance, RePaint~\cite{repaint} iteratively refines an inpainted region through a corruption-aware sampling loop, while DDRM~\cite{DDRM} adapts diffusion priors to tasks like denoising or super-resolution by enforcing consistency with the degraded observations. DreamClean~\cite{dreamclean} further extends unsupervised test-time sampling to remove unknown corruption. Most of these methods focus on local artifact correction or low-level enhancements. By contrast, our approach supports various geometric transformations and can synthesize new content guided by user prompts and masks.

\section{Method}
\label{sec:method}

\begin{figure}[t]
  \centering
  \includegraphics[width=\linewidth]{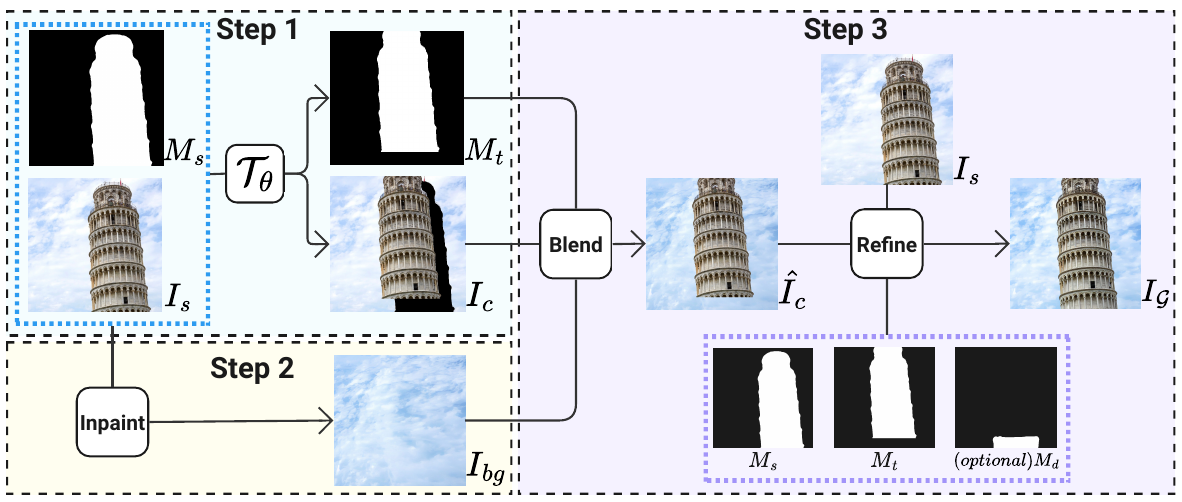}
  \caption{Overview of our geometric image editing pipeline.
  }
  \label{fig:main}
\end{figure}

\subsection{Geometric Editing Pipeline}

\noindent Our goal is to perform geometric editing on any object within the image, manipulating its shape, orientation, or position while preserving the overall coherence and realism of the scene. Our pipeline (Fig.~\ref{fig:main}) comprises \textbf{three} steps:

\vspace{1mm}\noindent\textbf{Step 1:~Object Transformation.}~The pipeline takes a source image $I_s$ containing an object to be edited, along with a binary mask $M_s$ indicating the source region for the object. To reduce the burden of drawing precise masks, we employ an interactive segmentation model~\cite{SAM} that generates $M_s$ with just a few user clicks. 
Next, we convert user's editing instruction (\eg, ``rotate the object along the $z$-axis by 30$^\circ$'') to a transformation function $\mathcal{T}_{\theta}$ with parameters $\theta$. The transformation function receives $I_s$ and $M_s$ as inputs, and generates transformed $M_t$ that indicates the target region for the object, and a {\em coarse} image $I_c$:
\[
I_c, M_t = \mathcal{T}_{\theta}(I_s, M_s).
\]
For purely 2D edits, $\mathcal{T}_{\theta}$ simply represents an affine transformation. For more advanced 3D edits, we first estimate scene depth using a depth estimator~\cite{DepthAnything}, then apply geometric transformations in 3D space and re-project the transformed object back into the image. 
For more demanding 3D changes, we employ single-image 3D lifting methods such as SV3D~\cite{SV3D,MAGICMAN} for a more complete 3D representation. 


\vspace{1mm}\noindent\textbf{Step 2:~Source Region Inpainting.}~With the object relocated, the original source region often requires cleanup (see $I_c$ in~\cref{fig:main}). This step aims to generate a clean background image $I_{bg}$ by inpainting the source region of $I_s$, ensuring natural blending with surrounding pixels:
\[  
I_{bg} = \text{\tt Inpaint}(I_s, M_s).
\]

\vspace{1mm}\noindent\textbf{Step 3:~Target Region Refinement.}~Given the target mask $M_t$ and the coarse image $I_c$ from Step 1, and $I_{bg}$ with clean background in the source object location obtained from Step 2, we can easily {\tt Blend} them together to create a composite image $\hat{I}_c$:
\[
\hat{I}_c = M_t \cdot I_c + (1-M_t)\cdot I_{bg}.
\]
$\hat{I}_c$ can be imperfect: as $I_c$ and $I_{bg}$ are separately built, the blended result may fall short in unnatural boundaries around the target object regions. 
More critically, occlusion or incompleteness of the original object can severely compromise the realism of the edited result. As an example in~\cref{fig:main}, the tower in $\hat{I}_c$ is up in the air without realistic structure in the base part. These limitations necessitate an additional refinement step on $\hat{I}_c$. 
This step requires many inputs obtained from previous steps:
\[
I_{\mathcal{G}} = \text{\tt Refine}(\hat{I}_c, I_s, M_s, M_t,[M_d]),
\]
where $M_d$ is an {\it optionally} user-provided completion mask for controlled content generation based on $\hat{I}_c$.

\begin{figure*}[tb]
  \centering
  \includegraphics[width=1.00\linewidth]{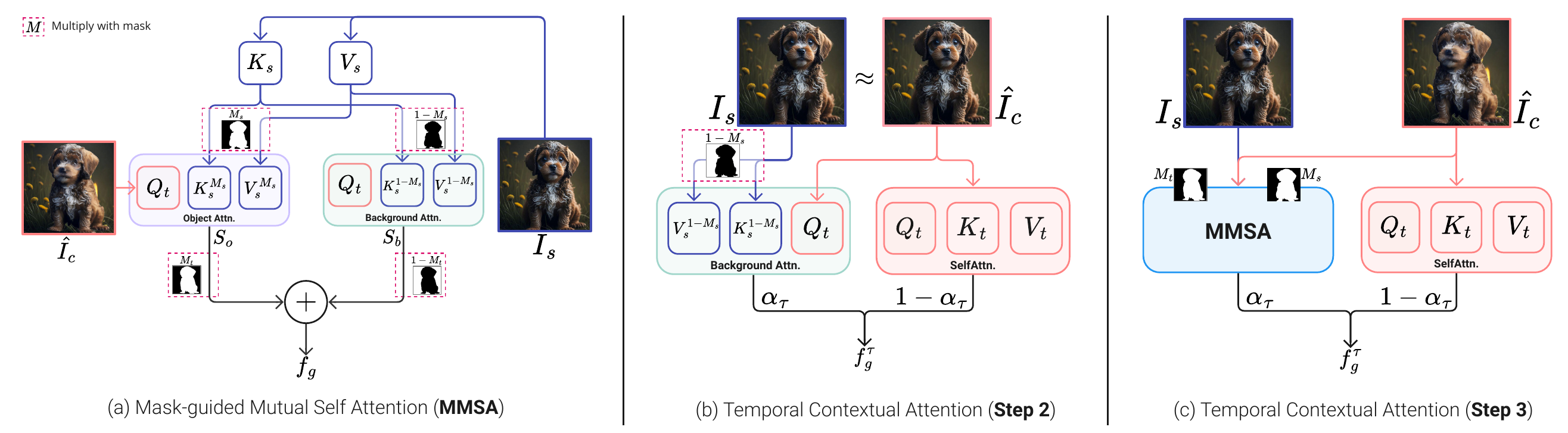}
\caption{Comparison of Context Aggregation Methods. This figure illustrates different approaches for context alignment in image editing tasks: (a) MMSA~\cite{MasaCtrl} replaces Key-Value (KV) pairs and enforces explicit feature interaction between regions. (b) TCA (Step2) for source region inpainting, and (c) TCA (Step3) for target region refinement, which smoothly transition
from MMSA to full self-attention.}
  \label{fig:Attention}
\end{figure*}

Although {\tt Inpaint} and {\tt Refine} have different objectives, both boil down to refining selected pixels using existing context from the available contexts. To that end, we propose a training-\textbf{free} image re\textbf{fine}ment approach, \textbf{\pipeline{}}, to perform these steps in a unified manner. In what follows, we detail the \pipeline{} framework and how it integrates into Steps~2 and~3.

\subsection{Training-Free Image Refinement}
A widely adopted approach for diffusion-based editing is to perform DDIM inversion on the input image and manipulate the latent at each denoising step. Here, we invert the source image $I_s$ \emph{once} and reuse its latent for both Step~2 (source region inpainting) and Step~3 (target region refinement). Once the composite image $\hat{I_c}$ is generated, we additionally invert it to initialize the latent representation for Step~3. Past studies~\cite{MasaCtrl,P2P,style_align,ConsiStory,fpe,Cross_image} demonstrate that manipulating self-attention features can be highly effective, given the rich semantic information learned by large-scale pretrained models. However, it is crucial to manage \emph{where} and \emph{how} such modifications occur. To this end, we introduce three complementary modules:
\begin{itemize}
    \item \textbf{\Attention{} (\attention{})} in~\cref{sec:tca}, which embodies the mechanism to smoothly transition from mask-guided mutual attention to full self-attention and preserve global structure during large edits.
    \item \textbf{Local Perturbation (LP)} in~\cref{sec:lp}, a method that selectively injects noise (via DDPM updates) into user-defined regions to permit substantial content changes without disturbing the rest of the image.
    \item \textbf{Content-specified Generation (CG)} in~\cref{sec:cg}, which utilizes classifier-free guidance and cross-attention based on user-provided prompts to steer newly generated content, ensuring desired details appear only where intended.
\end{itemize}
Together, these modules enable \pipeline{} to perform inpainting and refinement in a training-free manner while preserving the realism and coherence of the edited image.

\subsubsection{\Attention{}}
\label{sec:tca}
 
Diffusion-based editing aims to adjust latents in editable regions while keeping other parts intact. As self-attention layers capture feature dependencies, they enable diffusion models to have ``repairing abilities'': corrupting the regions to be edited and then performing the denoising process can harmonize corrupted regions and generate plausible results. But it is difficult to control both {\it where} such ``repairment'' occurs and its intensity. Mask-guided mutual self-attention (MMSA)~\cite{MasaCtrl} (illustrated in~\cref{fig:Attention}~(a)) avoids this by restricting each query in the self-attention layers to attend only to the corresponding masked features:
\begin{equation}
\label{eq:mtsa}
\begin{aligned}
    S_o &= \text{SelfAttn}(Q_{t}, K_{s}, V_{s};\, M_s), \\
    S_b &= \text{SelfAttn}(Q_{t}, K_{s}, V_{s};\, 1-M_s), \\
    f_g &= S_o \cdot M_t \;+\; S_b \cdot (1-M_t),
\end{aligned}
\end{equation}
where $S_o$ and $S_b$ represent the self-attention outputs for objects and backgrounds, respectively, and $f_g$ is the final feature output of the whole module. While effective, this approach is limited in generating structural content if large edits are desirable (shown in~\cref{sec:ablation}).

Our key observation (aligned with the literature~\cite{P2P,MasaCtrl,focusonins}) is that major content changes tend to occur at the beginning of the diffusion process, whereas later steps naturally refine local details. Hence, we seek a balanced approach that starts with MMSA to preserve global structure and then gradually incorporates self-attention as denoising progresses, enabling local adjustments without compromising the larger transformation.

To implement this idea, we propose a \emph{\Attention{} (\attention{})} mechanism shown in~\cref{fig:Attention} (c) that dynamically blends MMSA with self-attention from the target over the diffusion steps. Formally, let $S_{t}=\text{SelfAttn}(Q_{t}, K_{t}, V_{t})$ be the self-attention output capturing global context from the current image latent. We then define the final attention output $f_g^{\tau}$ at step $\tau$ as
\[
    f_g^{\tau} \;=\; (1 - \alpha_{\tau}) \, S_{t}
    \;+\; \alpha_{\tau} 
    \Bigl[
      S_{o} \cdot M_t \;+\; S_{b} \cdot (1 - M_t)
    \Bigr],
\]
where 
\[
    \alpha_{\tau} = \frac{\tau_1 - \tau}{\tau_1 - \tau_0}
\]
is a {\it temporally changing} blend weight that smoothly transitions from relying on MMSA at early steps (for large content edits) to incorporating self-attention at later steps (for fine details and completion). Concretely, $\alpha_{\tau}$ linearly decreases from $1$ when $\tau = \tau_0$ to $0$ when $\tau = \tau_1$, where $\tau_0$ is the initial denoising step, and $\tau_1$ is the final denoising step.
A straightforward alternative is to define a single threshold: for steps it, use MMSA from the source; once past it, switch entirely to self-attention. This approach is highly dependent on threshold choice and incurs varied performance for different thresholds (see Appendix for details). Our approach is free from this concern and thus more robust.

TCA can be applied to both Step~2 (\cref{fig:Attention}(b)) and Step~3 (\cref{fig:Attention}(c)), depending on the masks and the values of $\tau_0$ and $\tau_1$. For Step~2, we maintain background self-attention and combine it with full self-attention through $\alpha_{\tau}$, ensuring that source regions query only the background of $I_s$ while leveraging self-dependency for self-remedy. For Step~3, when structure completion is required, we extend the target mask $M_t$ by incorporating the user-defined mask $M_d$, forming the full target mask $M_{t}^* = M_t \cup M_d$. This allows the completion region to focus on the foreground region of $I_s$ while smoothly combining self-attention for coherent content completion.

\subsubsection{Local Perturbation}
\label{sec:lp}
While DDIM~\cite{DDIM} provides deterministic denoising updates, in many editing scenarios there is a need to \emph{unfreeze} certain regions to allow more dramatic changes, such as inpainting large occlusions or fixing problematic artifacts. To accomplish this, we introduce \emph{Local Perturbation} (LP), which injects controlled stochasticity \emph{only} within a specified region $\mathcal{M}$. Formally, we selectively apply a DDPM-like update in $\mathcal{M}$ while retaining DDIM updates elsewhere:
\[
x_{t-1} =
\begin{cases}
\mathrm{DDPM}(x_t), & \text{if } x \in \mathcal{M},\\
\mathrm{DDIM}(x_t), & \text{otherwise}.
\end{cases}
\]
By applying stochasticity only in $\mathcal{M}$, LP provides extra flexibility to rearrange or complete local structures within the region without disrupting content outside it. 
For Step~2, $\mathcal{M}$ typically corresponds to the source mask $M_s$, while in Step~3, $\mathcal{M}$ is defined by the user-defined structure completion mask $M_d$  or other regions requiring substantial modification, such as areas containing inpainting artifacts or the boundaries between objects and the background. More details can be found in the Appendix. 

\subsubsection{Content-specified Generation}
\label{sec:cg}
Though LP encourages greater variety in how masked regions evolve, it still relies on the model’s generative prior to fill in missing details, which can be arbitrary if the desired content is not specified. To address this, we introduce a textual conditional input $\mathcal{C}$ (\eg, ``add a missing foot'', ``make an empty scene'') and define two regions: $\mathcal{M}_1$ for localized cross-attention and $\mathcal{M}_2$ to guide classifier-free guidance~\cite{classifierfreeguidance}. 

Within $\mathcal{M}_1$, we replace the source key--value pairs with those projected from the textual embeddings $(K_{\mathcal{C}}, V_{\mathcal{C}})$:

\begin{equation}
\label{eq:CGCrossAttn}
\resizebox{0.7\hsize}{!}{$
\begin{aligned}
\tilde{f}_{t} \;=\; & \mathrm{CrossAttn}(Q_{t}, K_{\mathcal{C}}, V_{\mathcal{C}})\,\cdot\,\mathcal{M}_1 \;+\; \\
& \mathrm{CrossAttn}(Q_{t}, K_{\varnothing}, V_{\varnothing})\,\cdot\,(1 - \mathcal{M}_1),
\end{aligned}
$}
\end{equation}
where $Q_{t}$ is the target query and $(K_{\varnothing}, V_{\varnothing})$ are  text embeddings from null texts. This ensures cross-attention is locally focused on $\mathcal{M}_1$, enabling precise, user-specified generation within the region while preserving external content.

Furthermore, we apply classifier-free guidance~\emph{exclusively} $\mathcal{M}_2$:
\begin{equation}
\label{eq:cfg}
\resizebox{0.9\hsize}{!}{$
\hat{\epsilon}_\theta(x_t,\mathcal{C})
=\;
\epsilon_\theta(x_t,\varnothing)
\;+\;
w\,\Bigl[\epsilon_\theta(x_t,\mathcal{C})\;-\;\epsilon_\theta(x_t,\varnothing)\Bigr]\cdot \mathcal{M}_2,
$}
\end{equation}
where $w$ is the guidance scale. This confines semantic directives in $\mathcal{C}$ to $\mathcal{M}_2$, preventing unintended changes to well-established parts of the image. For Step~2, $\mathcal{M}_1 = \mathcal{M}_2 = M_{s}$, where $M_{s}$ is the source region mask. For Step~3, $\mathcal{M}_1 = M_t^*$, while $\mathcal{M}_2 = M_d$. 
By combining LP’s controlled stochasticity with CG’s text-driven conditioning, users can refine or synthesize content exactly where needed while preserving overall image coherence.

\section{Experiments}

\subsection{Experimental Settings}
\label{sec:experimental_setting}
Additional implementation details of our method, compared approaches, dataset, and metric calculations are provided in the Appendix.

\vspace{1mm} \noindent \textbf{Implementation Details.}  
We adopt Stable Diffusion v1-5~\cite{SD} as our base model, with an image resolution of $512 \times 512$, consistent with previous methods. 
The number of the denoising steps $\tau_1$ is set to 50. \attention{} is applied at all timesteps, starting from the tenth U-Net layer, following ~\cite{MasaCtrl}.
For Step~2, we set $\tau_0$ to 1, minimizing the influence of residual object features on background reconstruction. For Step~3, $\tau_0$ is set to 13 to balance structural completion and detail preservation. For general refinement without structural completion, $\tau_0$ is set to 25 for fine-grained adjustments. CG is applied in both Step~2 and Step~3 with a default guidance scale $w=7.5$. 

\vspace{1mm} \noindent \textbf{Datasets.}~To evaluate on geometric editing, we construct a comprehensive benchmark,~\textbf{GeoBench}, by combining source images from PIE-Bench~\cite{pie-BENCH} and Subjects200K~\cite{Subjects200k}, which contain a mix of real and synthetic images with apparent objects suitable for this task. For each source image, we provide multiple geometric editing instructions, including object-centric operations such as move, rotate, and resize, each with varying directions and three intensity levels (\ie, easy, medium, hard). The GeoBench dataset comprises 811 source images and 5,988 editing instructions in total, offering a robust foundation for evaluating geometric editing methods. Our benchmark includes diverse editing scenarios and a challenging subset requiring {\it structural completion}. Additionally, we provide detailed captions and object category labels for all images, as well as annotated regions for structural completion tasks. 

\vspace{1mm}\noindent \textbf{Metrics.}~We employed seven metrics to quantitatively evaluate the generated results from three different perspectives. (1) {\bf Image Quality}. FID~\cite{FID} is computed to comprehensively evaluate the quality of the generated images. We randomly sample 2k images from PIE-Bench~\cite{pie-BENCH} and Subjects200K~\cite{Subjects200k} as data from the target. In addition, we separately use Kernel distances~\cite{KD} (KD) and DINOv2~\cite{DINOv2} feature distance (DINOv2) to improve the FID and obtain more comprehensive results. (2) {\bf Consistency}. Inspired by VBench~\cite{vbench}, we adopt Subject Consistency (SUBC) and Background Consistency (BC) to assess the fidelity of the generated image to the input source images. After separating the subject from the background using $M_{s}$ and $M_{t}$, we calculate the cosine similarity between their foregrounds and between their backgrounds in the feature space. (3) {\bf Editing Effectiveness}. In image editing tasks, it is crucial to evaluate whether the generated images adhere to the input editing instructions. We employ the same Warp Error (WE) and Mean Distance(MD) as GeoDiffuser~\cite{geodiffuser} to measure editing effectiveness, which warps the source object to the target location and then computes L1 error within masked regions of the generated images.

\vspace{1mm}\noindent \textbf{Baselines.}~Our evaluation includes two main aspects: (1) direct comparisons with state-of-the-art image editing methods, and (2) comparisons with representative inpainting methods integrated into our framework to address Step~2 and Step~3. For (1), we compare with DragonDiffusion~\cite{Dragon}, Self-Guidance~\cite{Self-Guidance}, Motion-Guidance~\cite{MotionGuidance}, and RegionDrag~\cite{RegionDrag}, DragDiffusion~\cite{dragdiffusion}, GeoDiffuser~\cite{geodiffuser}, DiffusionHandles~\cite{Diffusionhandles}, and DesignEdit~\cite{DesignEdit}. All methods are implemented based on their official codebases, with minimal adjustments to fit our benchmark~(see Appendix). 
For (2), we compare with BrushNet~\cite{Brush}, Stable Diffusion Inpainting~\cite{stableinpainting}, LaMa~\cite{Lama}, and MAT~\cite{MAT}. Since LaMa~\cite{Lama} and MAT~\cite{MAT} learned to remove content (inpainting) rather than performing target region refinement, they are excluded from the main comparison but included in our user study (see Section~\ref{sec:user_study}).

\subsection{Comparison with Other Methods} 
\begin{table*}[tb]
\footnotesize
\setlength{\tabcolsep}{1.5mm}
\caption{Comparison on 2D Edits, 3D Edits, and Structure Completion (SC) tasks. Best results are in bold, second best in underlined.}
\label{tab:comparison}
\renewcommand\arraystretch{1.0}
\centering
\begin{tabular}{@{}l|l|c|ccccccc@{}}
\toprule
Methods & External Model & Editing Type & FID & DINOv2 & KD & SUBC & BC & WE & MD \\
\midrule
Self-Guidance~\cite{Self-Guidance} & SAM~\cite{SAM} & \multirow{9}{*}{2D} & 49.15 & 647.56 & 0.438 & 0.575 & 0.759 & 0.268 & 116.89 \\
RegionDrag~\cite{RegionDrag} & SAM~\cite{SAM} &  & 40.21 & 504.50 & 0.241 & 0.796 & \underline{0.970} & 0.120 & 32.50 \\
DragonDiffusion~\cite{Dragon} & SAM~\cite{SAM} &  & 37.09 & 507.67 & \underline{0.144} & 0.840 & 0.968 & 0.158 & 32.36 \\
MotionGuidance~\cite{MotionGuidance} & SAM~\cite{SAM}, RAFT~\cite{RAFT} &  & 106.39 & 1189.23 & 3.871 & 0.521 & 0.736 & 0.186 & 90.03 \\
DragDiffusion~\cite{dragdiffusion} & SAM~\cite{SAM} &  & 36.58 & \underline{455.68} & \textbf{0.142} & 0.758 & 0.966 & 0.199 & 41.31 \\
Diffusion Handles~\cite{Diffusionhandles} & SAM~\cite{SAM}, LaMa~\cite{Lama}, DepthAnything~\cite{DepthAnything} &  & 44.81 & 549.69 & 0.618 & 0.725 & 0.852 & 0.180 & 40.27 \\
GeoDiffuser~\cite{geodiffuser} & SAM~\cite{SAM}, DepthAnything~\cite{DepthAnything} &  & \textbf{33.89} & \textbf{437.75} & 0.173 & 0.762 & 0.938 & 0.166 & 34.88 \\
DesignEdit~\cite{DesignEdit} & SAM~\cite{SAM} &  & 35.22 & 480.91 & 0.179 & \underline{0.874} & 0.959 & \underline{0.098} & \underline{10.15} \\
\rowcolor{gray!10}
{\bf FreeFine} & SAM~\cite{SAM} &  & \underline{34.72} & 478.18 & \underline{0.144} & \textbf{0.907} & \textbf{0.971} & \textbf{0.055} & \textbf{9.25} \\
\bottomrule

DragDiffusion~\cite{dragdiffusion} & SAM~\cite{SAM} & \multirow{4}{*}{3D} & 157.42 & \textbf{1867.02} & \underline{0.348} & 0.603 & \textbf{0.958} & 0.199 & 61.97 \\
Diffusion Handles~\cite{Diffusionhandles} & SAM~\cite{SAM}, LaMa~\cite{Lama}, DepthAnything~\cite{DepthAnything} &  & 156.90 & 1882.66 & 0.523 & 0.705 & 0.882 & 0.128 & \underline{26.10} \\
GeoDiffuser~\cite{geodiffuser} & SAM~\cite{SAM}, DepthAnything~\cite{DepthAnything} &  & 152.06 & 1894.26 & 0.351 & \underline{0.749} & 0.941 & \underline{0.097} & 34.34 \\
\rowcolor{gray!10}
{\bf FreeFine} & SAM~\cite{SAM}, DepthAnything~\cite{DepthAnything} &  & \textbf{150.89} & \underline{1879.69} & \textbf{0.310} & \textbf{0.786} & \underline{0.956} & \textbf{0.079} & \textbf{20.32} \\
\bottomrule

BrushNet~\cite{Brush} & SAM~\cite{SAM} & \multirow{3}{*}{SC} & \underline{186.93} & \textbf{2516.52} & \textbf{0.971} & \underline{0.925} & \underline{0.948} & \underline{0.060} & \underline{11.31} \\
SD-inpainting~\cite{stableinpainting} & SAM~\cite{SAM} &  & 193.71 & 2556.44 & 1.047 & 0.913 & 0.928 & 0.064 & 14.43 \\
\rowcolor{gray!10}
{\bf FreeFine} & SAM~\cite{SAM} &  & \textbf{184.84} & \underline{2526.38} & \underline{0.982} & \textbf{0.928} & \textbf{0.952} & \textbf{0.056} & \textbf{9.56} \\
\bottomrule

\end{tabular}
\vspace{-10pt}
\end{table*}
\begin{figure*}[tb]
  \centering
  \includegraphics[width=\linewidth]{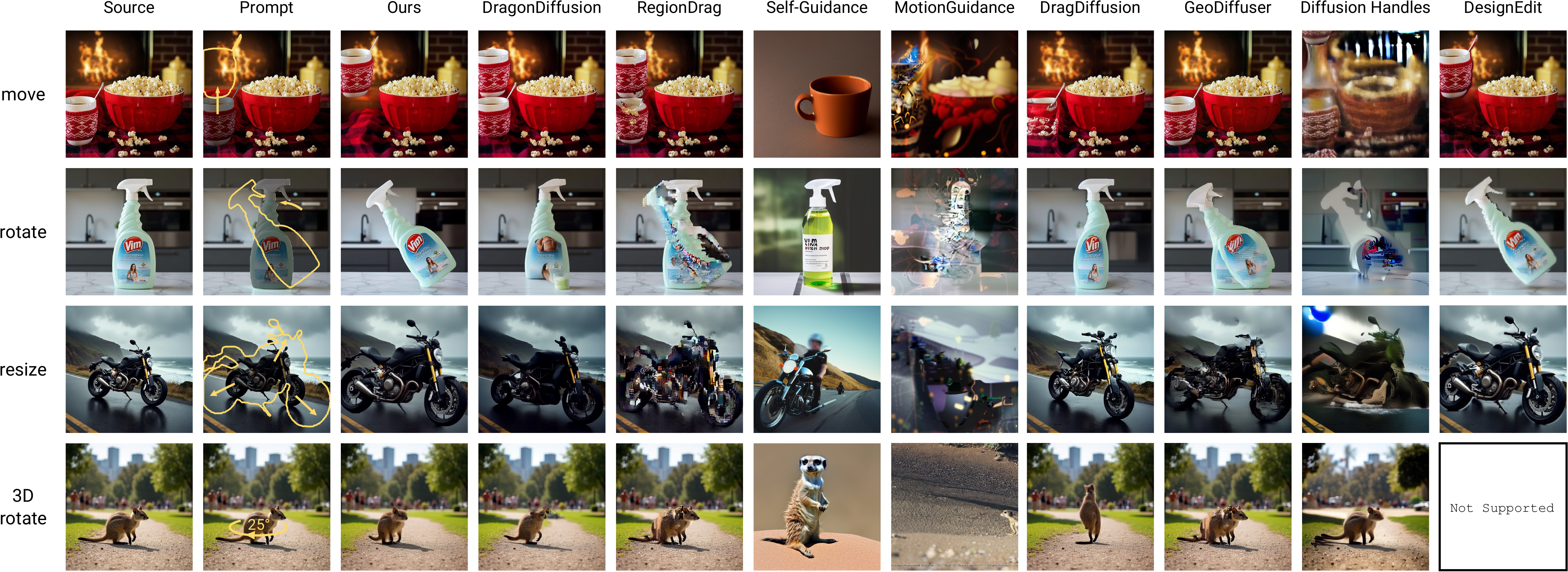}
  \caption{Qualitative comparison with state-of-the-art editing approach. {\bf Zoom-in} for better details. Since DesignEdit~\cite{DesignEdit} is limited to 2D layout-based editing and lacks 3D task support, its 3D editing result is marked with a "Not Supported" placeholder. Additional results are supplemented in the Appendix. }
  \label{fig:compare_edit}
\end{figure*}

\noindent {\bf Quantitative Results.}~We conduct a comprehensive quantitative evaluation of FreeFine against SotA
methods across 2D edits, 3D edits, and structure completion tasks in Table \ref{tab:comparison}. FreeFine demonstrates consistent superiority across all scenarios and metrics:
For 2D edits, FreeFine outperforms all the counterparts, with SUBC, WE and MD significantly better. Among compared methods, DesignEdit and GeoDiffuser stand out, where the former excels at edit precision and the latter is better at image quality.
For 3D edits, methods (\eg, DesignEdit~\cite{DesignEdit}, RegionDrag\cite{RegionDrag}) good at 2D edits fail to support holistic object 3D rotation by design. FreeFine undoubtedly achieves the best performance across all evaluation dimensions under the same depth-based transformation paradigm.
For structure completion tasks, we further benchmark against representative inpainting models by integrating them into our editing pipeline. Our method comprehensively outperforms training-based methods like BrushNet, with no additional training needed.
\vspace{1mm} \noindent \textbf{Qualitative Results.}~
We present a qualitative evaluation of our method against baseline approaches, focusing on both geometric editing and structural completion tasks. As illustrated in Fig.~\ref{fig:compare_edit}, our method achieves high-fidelity editing without noticeable artifacts and enables more accurate and diverse transformations. All other methods have clear artifacts: DragonDiffusion and RegionDrag leave residual artifacts in the Teacup example and fail in the rotate and resize examples; SelfGuidance, MotionGuidance and DiffusionHandles change the image contents significantly, explaining their lower SUBC and BC in~\cref{tab:comparison}. GeoDiffuser also faces residual artifacts and struggles in move example. DesignEdit, though excels at precision, falls short in the realism of edited results, corresponding to its relatively lower image quality in~\cref{tab:comparison}. Our method is the most successful approach in the 3D rotation example, while all other methods fail in this case.

\begin{figure*}[h!]
  \centering
  \includegraphics[width=0.95\linewidth]{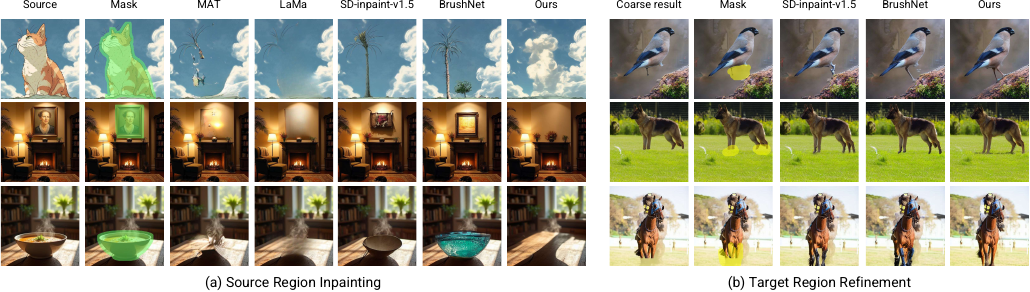}
  \caption{Qualitative comparison with state-of-the-art inpainting methods. Notably, our method is \textbf{training-free}, while all the compared methods are training-based.}
  \label{fig:compare_inpaint}
\end{figure*}

We further compare our method with representative inpainting models in two key aspects: (a) Source Region Inpainting and (b) Target Region Refinement, as shown in Fig.~\ref{fig:compare_inpaint}. For (a), our method achieves less hallucination compared to BrushNet and SD-inpainting, while producing more realistic textures and finer details than LaMa and MAT. For (b), our method not only generates complex structures while maintaining contextual consistency with the object (\eg, the bird and the horse) and producing more details (\eg, dog's shadow).


\vspace{1mm} \noindent \textbf{User Study.}
\label{sec:user_study}
For a comprehensive quantitative evaluation, we conducted a user study to assess the perceptual quality and editing effectiveness of our method. We recruited 35 participants with diverse backgrounds in computer vision and collected 2,622 valid votes. Each participant was presented with 30 randomly selected editing samples from different tasks (2D-edits, 3D-edits, region refinement and region inpainting). Each sample contained the original image along with a series of corresponding edited results generated by our method and other comparative models. Participants were asked to pick the best edited images based on: 1) {\bf Image Quality}: How visually realistic and artifact-free the edited image appears; 2) {\bf Consistency}: How well the edited image preserves the original subject and background; 3) {\bf Editing Effectiveness}: How accurately the edited image reflects the intended geometric transformations (e.g., move, rotate, or resize). In the Appendix, we show our user study has close alignment with our quantitative metrics in~\cref{sec:experimental_setting}.

\begin{figure}[h!]
  \centering
  \includegraphics[width=0.95\linewidth]{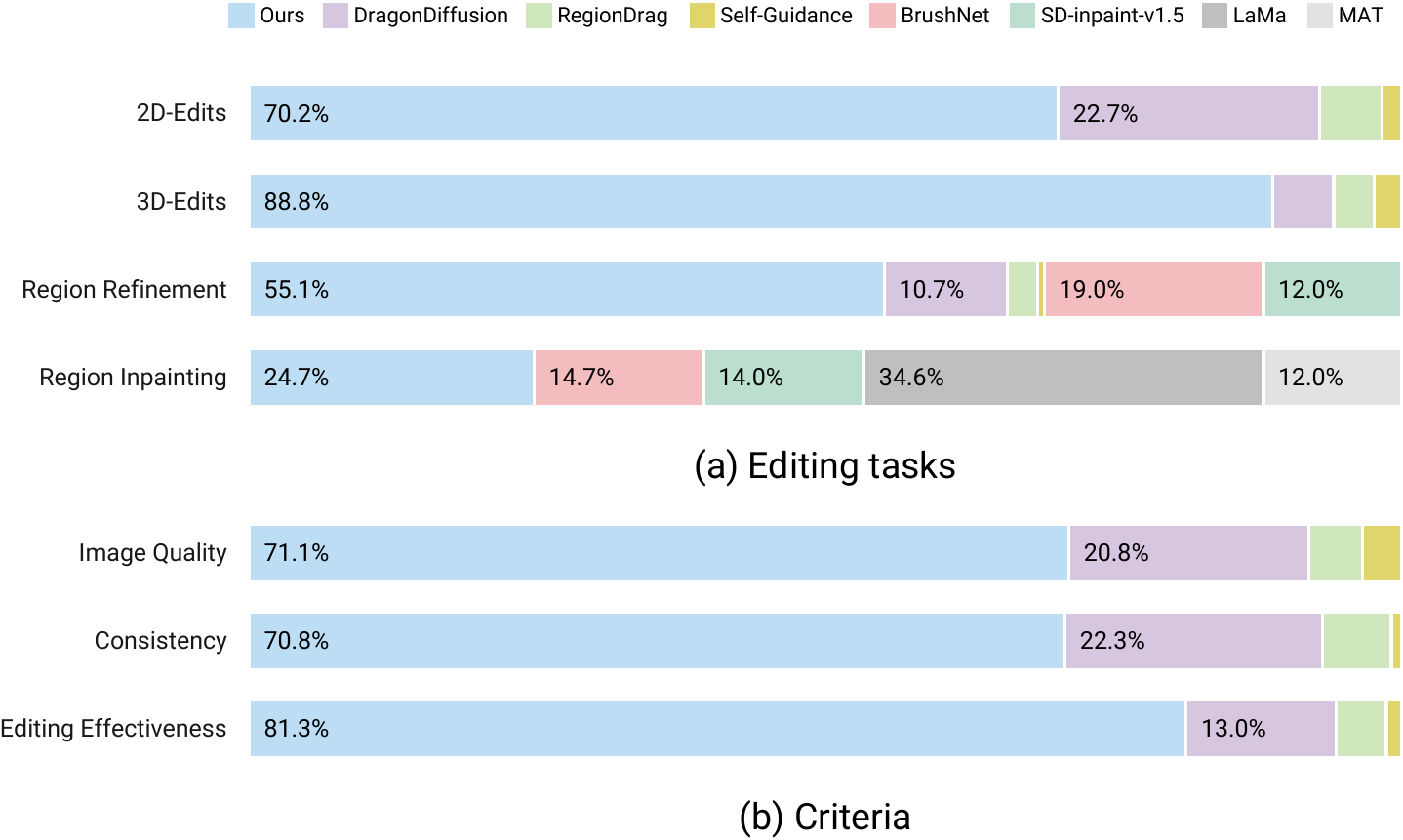}
  \caption{Visualization results of the user study. Participants preferred our edited images both in the different editing tasks and from three different criteria.}
  \label{fig:user_study}
  \vspace{-5pt}
\end{figure}

As shown in Fig.~\ref{fig:user_study} (a), our method performs well in all editing tasks, especially 2D-edits (70.2\% user preference) and 3D-edits  (88.8\% user preference). 
Notably, while our method leads in most tasks, it is second to LaMa~\cite{Lama} in region inpainting, which is better at removing objects but tends to leave a blurry mess. As for the evaluation criteria in 2D-edits and 3D-edits (Fig.~\ref{fig:user_study} (b)), our method is significantly better: the voting rates are all more than three times ahead. 
Despite DragonDiffusion~\cite{Dragon}’s comparable FID scores to ours in~\cref{tab:comparison}, it lags greatly in perceptual quality assessments (20.8\% vs. 71.1\%). These findings further validate the robustness and practicality of our method in real-world geometric editing scenarios. More details about the user study and the statistics are included in the Appendix.

\subsection{Ablation Studies} 
\label{sec:ablation}

\begin{figure*}[h!]
  \centering
  \includegraphics[width=0.95\linewidth]{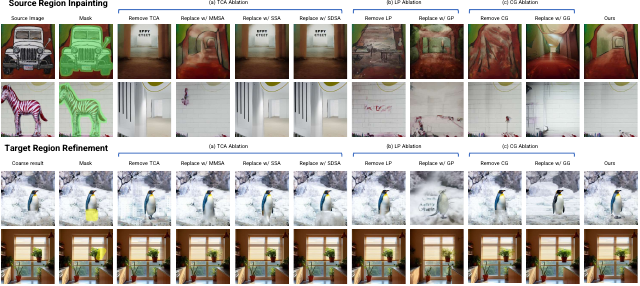}
\caption{Ablation studies on the impact of removing individual components from \textbf{FreeFine} and different internal variations of each component while keeping other techniques applied at the same scale.}
  \label{fig:ablation}
\end{figure*}

We conduct ablation studies on \attention{}, LP, and CG to dissect the impact of each component on the generated results. While ablating one component, we make sure other techniques are kept the same. 


\vspace{1mm}\noindent \textbf{Attention Mechanisms.}~Beside MMSA~\cite{MasaCtrl}, we compare \attention{} with two additional attention mechanisms: (a) Shared Self-Attention (SSA)~\cite{style_align}, which shares a set of Key-Value (KV) pairs encoded from different text prompts to encourage the model to align features implicitly across the entire latent space (b) Subject-Driven Self-Attention (SDSA)~\cite{ConsiStory}, which selectively applies this sharing selectively only on the key and value vectors of foreground regions for object alignment. In our context, the shared KV pairs are constructed by concatenating the keys from the source and target $[K_s, K_t]$, and similarly for the values to form $[V_s, V_t]$.
As shown in Fig.~\ref{fig:ablation}, for source region inpainting,
SSA, SDSA, and standard Self-Attention (removing \attention{}) produce similar results where undesired changes appear in the source region, due to the non-constrained global feature sharing.
MMSA explicitly restricts attention between the target region and the background region, achieving object removal but introducing texture artifacts (\eg, stains on the wall in the second row). 
In contrast, due to the repair of the self-attention in larger denoisng timesteps, \attention{} generates better background. For target region refinement examples, standard Self-Attention alters details in regions where changes are not desired. Both SSA and SDSA do not incorporate masks, therefore showing limited completion capabilities. MMSA struggles when $M_d$ contains semantics not present in $M_s$ (\eg, the penguin's feet), whereas \attention{} demonstrates smooth and robust completion performance, ensuring contextual consistency.

\vspace{1mm} \noindent \textbf{Perturbation and Guidance.} We study the impact of LP and CG for both source region inpainting and target region refinement. Removing LP is equivalent to using DDIM deterministic denoising, whereas the other option is to use DDPM stochastic denoising, which we refer to as Global Perturbation (GP). Removing CG is equivalent to setting $\mathcal{C}=\varnothing$ in~\cref{eq:CGCrossAttn} and~\cref{eq:cfg}. Setting $\mathcal{M}_1=1$ in~\cref{eq:CGCrossAttn} and $\mathcal{M}_2=1$ in~\cref{eq:cfg} leads to Global Guidance (GG).
As illustrated in Fig.~\ref{fig:ablation}, for both source region inpainting and target region refinement, removing LP reduces randomness in results while using GP leads to undesired global changes.
Similarly, removing CG eliminates explicit context guidance, hindering the generation of desired content. Replacing CG with global attention also alters textures across the entire image, reducing background and subject consistency.

\section{Conclusion and Limitations}
\vspace{-2mm}
We present a principled framework for geometric image editing that systematically addresses the subtasks of object transformation, source region inpainting, and target region refinement. By decoupling these tasks, our approach more effectively balances large structural changes against fine-grained adjustments. The proposed \pipeline{}, equipped with {Temporally Contextual Attention}, {Local Perturbation}, and {Content-specified Generation}, demonstrates consistent gains in both fidelity and edit precision on our proposed {GeoBench} benchmark.
A detailed discussion of limitations is provided in the Appendix. We believe that continued efforts in either addressing these limitations or developing more powerful and intuitive geometric image editing methods could benefit both the research community and industrial applications.

\section*{Acknowledgments}

This work was supported by the NSFC
(62441615 and 62225603).

{
    \small
    \bibliographystyle{ieeenat_fullname}
    \bibliography{main}

\begin{thebibliography}{81}
\providecommand{\natexlab}[1]{#1}
\providecommand{\url}[1]{\texttt{#1}}
\expandafter\ifx\csname urlstyle\endcsname\relax
  \providecommand{\doi}[1]{doi: #1}\else
  \providecommand{\doi}{doi: \begingroup \urlstyle{rm}\Url}\fi

\bibitem[Alaluf et~al.(2024)Alaluf, Garibi, Patashnik, Averbuch-Elor, and Cohen-Or]{Cross_image}
Yuval Alaluf, Daniel Garibi, Or Patashnik, Hadar Averbuch-Elor, and Daniel Cohen-Or.
\newblock Cross-image attention for zero-shot appearance transfer.
\newblock In \emph{ACM SIGGRAPH}, pages 1--12, 2024.

\bibitem[Betker et~al.(2023)Betker, Goh, Jing, Brooks, Wang, Li, Ouyang, Zhuang, Lee, Guo, et~al.]{dalle3}
James Betker, Gabriel Goh, Li Jing, Tim Brooks, Jianfeng Wang, Linjie Li, Long Ouyang, Juntang Zhuang, Joyce Lee, Yufei Guo, et~al.
\newblock Improving image generation with better captions.
\newblock \emph{Computer Science. https://cdn. openai. com/papers/dall-e-3. pdf}, 2\penalty0 (3):\penalty0 8, 2023.

\bibitem[Binkowski et~al.(2018)Binkowski, Sutherland, Arbel, and Gretton]{KD}
Mikolaj Binkowski, Danica~J. Sutherland, Michael Arbel, and Arthur Gretton.
\newblock Demystifying {MMD} gans.
\newblock In \emph{ICLR}. OpenReview.net, 2018.

\bibitem[Bradski(2000)]{opencv}
G. Bradski.
\newblock {The OpenCV Library}.
\newblock \emph{Dr. Dobb's Journal of Software Tools}, 2000.

\bibitem[Cao et~al.(2023)Cao, Wang, Qi, Shan, Qie, and Zheng]{MasaCtrl}
Mingdeng Cao, Xintao Wang, Zhongang Qi, Ying Shan, Xiaohu Qie, and Yinqiang Zheng.
\newblock Masactrl: Tuning-free mutual self-attention control for consistent image synthesis and editing.
\newblock In \emph{ICCV}, pages 22503--22513, 2023.

\bibitem[Caron et~al.(2021)Caron, Touvron, Misra, J{\'{e}}gou, Mairal, Bojanowski, and Joulin]{DINO}
Mathilde Caron, Hugo Touvron, Ishan Misra, Herv{\'{e}} J{\'{e}}gou, Julien Mairal, Piotr Bojanowski, and Armand Joulin.
\newblock Emerging properties in self-supervised vision transformers.
\newblock In \emph{ICCV}, pages 9630--9640, 2021.

\bibitem[Chen et~al.(2024)Chen, Li, Dong, Zhang, He, Wang, Zhao, and Lin]{SHARE4V}
Lin Chen, Jinsong Li, Xiaoyi Dong, Pan Zhang, Conghui He, Jiaqi Wang, Feng Zhao, and Dahua Lin.
\newblock Sharegpt4v: Improving large multi-modal models with better captions.
\newblock In \emph{ECCV}, pages 370--387, 2024.

\bibitem[Cui et~al.(2021)Cui, McKee, and Lazebnik]{DressInOrder}
Aiyu Cui, Daniel McKee, and Svetlana Lazebnik.
\newblock Dressing in order: Recurrent person image generation for pose transfer, virtual try-on and outfit editing.
\newblock In \emph{ICCV}, pages 14618--14627. {IEEE}, 2021.

\bibitem[Cui et~al.(2024)Cui, Zhao, Zhang, Cao, Ma, and Wang]{stabledrag}
Yutao Cui, Xiaotong Zhao, Guozhen Zhang, Shengming Cao, Kai Ma, and Limin Wang.
\newblock Stabledrag: Stable dragging for point-based image editing.
\newblock In \emph{ECCV}, pages 340--356, 2024.

\bibitem[DeTone et~al.(2016)DeTone, Malisiewicz, and Rabinovich]{DeepHomogrophy}
Daniel DeTone, Tomasz Malisiewicz, and Andrew Rabinovich.
\newblock Deep image homography estimation.
\newblock \emph{CoRR}, abs/1606.03798, 2016.

\bibitem[Epstein et~al.(2023)Epstein, Jabri, Poole, Efros, and Holynski]{Self-Guidance}
Dave Epstein, Allan Jabri, Ben Poole, Alexei~A. Efros, and Aleksander Holynski.
\newblock Diffusion self-guidance for controllable image generation.
\newblock In \emph{NeurIPS}, 2023.

\bibitem[Fei et~al.(2023)Fei, Lyu, Pan, Zhang, Yang, Luo, Zhang, and Dai]{GenerativeDiffusionPrior}
Ben Fei, Zhaoyang Lyu, Liang Pan, Junzhe Zhang, Weidong Yang, Tianyue Luo, Bo Zhang, and Bo Dai.
\newblock Generative diffusion prior for unified image restoration and enhancement.
\newblock In \emph{CVPR}, pages 9935--9946, 2023.

\bibitem[Geng and Owens(2024)]{MotionGuidance}
Daniel Geng and Andrew Owens.
\newblock Motion guidance: Diffusion-based image editing with differentiable motion estimators.
\newblock In \emph{ICLR}, 2024.

\bibitem[Guo and Lin(2024)]{focusonins}
Qin Guo and Tianwei Lin.
\newblock Focus on your instruction: Fine-grained and multi-instruction image editing by attention modulation.
\newblock In \emph{CVPR}, pages 6986--6996, 2024.

\bibitem[Han et~al.(2018)Han, Wu, Wu, Yu, and Davis]{VITON}
Xintong Han, Zuxuan Wu, Zhe Wu, Ruichi Yu, and Larry~S. Davis.
\newblock {VITON:} an image-based virtual try-on network.
\newblock In \emph{CVPR}, pages 7543--7552. Computer Vision Foundation / {IEEE} Computer Society, 2018.

\bibitem[He et~al.(2025)He, Wu, Li, Kang, Zhang, Ye, Chen, Gao, Zhang, and Zhuang]{MAGICMAN}
Xu He, Zhiyong Wu, Xiaoyu Li, Di Kang, Chaopeng Zhang, Jiangnan Ye, Liyang Chen, Xiangjun Gao, Han Zhang, and Haolin Zhuang.
\newblock Magicman: Generative novel view synthesis of humans with 3d-aware diffusion and iterative refinement.
\newblock In \emph{AAAI}, pages 3437--3445, 2025.

\bibitem[Hertz et~al.(2023)Hertz, Mokady, Tenenbaum, Aberman, Pritch, and Cohen{-}Or]{P2P}
Amir Hertz, Ron Mokady, Jay Tenenbaum, Kfir Aberman, Yael Pritch, and Daniel Cohen{-}Or.
\newblock Prompt-to-prompt image editing with cross-attention control.
\newblock In \emph{ICLR}, 2023.

\bibitem[Hertz et~al.(2024)Hertz, Voynov, Fruchter, and Cohen{-}Or]{style_align}
Amir Hertz, Andrey Voynov, Shlomi Fruchter, and Daniel Cohen{-}Or.
\newblock Style aligned image generation via shared attention.
\newblock In \emph{CVPR}, pages 4775--4785, 2024.

\bibitem[Heusel et~al.(2017)Heusel, Ramsauer, Unterthiner, Nessler, and Hochreiter]{FID}
Martin Heusel, Hubert Ramsauer, Thomas Unterthiner, Bernhard Nessler, and Sepp Hochreiter.
\newblock Gans trained by a two time-scale update rule converge to a local nash equilibrium.
\newblock In \emph{NeurIPS}, pages 6626--6637, 2017.

\bibitem[Ho and Salimans(2022)]{classifierfreeguidance}
Jonathan Ho and Tim Salimans.
\newblock Classifier-free diffusion guidance.
\newblock \emph{CoRR}, abs/2207.12598, 2022.

\bibitem[Ho et~al.(2020)Ho, Jain, and Abbeel]{DDPM}
Jonathan Ho, Ajay Jain, and Pieter Abbeel.
\newblock Denoising diffusion probabilistic models.
\newblock In \emph{NeurIPS}, 2020.

\bibitem[Huang et~al.(2024)Huang, He, Yu, Zhang, Si, Jiang, Zhang, Wu, Jin, Chanpaisit, Wang, Chen, Wang, Lin, Qiao, and Liu]{vbench}
Ziqi Huang, Yinan He, Jiashuo Yu, Fan Zhang, Chenyang Si, Yuming Jiang, Yuanhan Zhang, Tianxing Wu, Qingyang Jin, Nattapol Chanpaisit, Yaohui Wang, Xinyuan Chen, Limin Wang, Dahua Lin, Yu Qiao, and Ziwei Liu.
\newblock {VBench}: Comprehensive benchmark suite for video generative models.
\newblock In \emph{CVPR}, 2024.

\bibitem[Jaderberg et~al.(2015)Jaderberg, Simonyan, Zisserman, and kavukcuoglu]{SpatialTransformer}
Max Jaderberg, Karen Simonyan, Andrew Zisserman, and koray kavukcuoglu.
\newblock Spatial transformer networks.
\newblock In \emph{NeurIPS}. Curran Associates, Inc., 2015.

\bibitem[Jia et~al.(2024)Jia, Yuan, Cheng, Wang, Li, Jia, and Zhang]{DesignEdit}
Yueru Jia, Yuhui Yuan, Aosong Cheng, Chuke Wang, Ji Li, Huizhu Jia, and Shanghang Zhang.
\newblock Designedit: Multi-layered latent decomposition and fusion for unified {\&} accurate image editing.
\newblock \emph{CoRR}, abs/2403.14487, 2024.

\bibitem[Ju et~al.(2024{\natexlab{a}})Ju, Liu, Wang, Bian, Shan, and Xu]{Brush}
Xuan Ju, Xian Liu, Xintao Wang, Yuxuan Bian, Ying Shan, and Qiang Xu.
\newblock Brushnet: {A} plug-and-play image inpainting model with decomposed dual-branch diffusion.
\newblock In \emph{ECCV}, pages 150--168, 2024{\natexlab{a}}.

\bibitem[Ju et~al.(2024{\natexlab{b}})Ju, Zeng, Bian, Liu, and Xu]{pie-BENCH}
Xuan Ju, Ailing Zeng, Yuxuan Bian, Shaoteng Liu, and Qiang Xu.
\newblock Pnp inversion: Boosting diffusion-based editing with 3 lines of code.
\newblock \emph{International Conference on Learning Representations ({ICLR})}, 2024{\natexlab{b}}.

\bibitem[Kawar et~al.(2022)Kawar, Elad, Ermon, and Song]{DDRM}
Bahjat Kawar, Michael Elad, Stefano Ermon, and Jiaming Song.
\newblock Denoising diffusion restoration models.
\newblock In \emph{NeurIPS}, 2022.

\bibitem[Kerbl et~al.(2023)Kerbl, Kopanas, Leimk{\"{u}}hler, and Drettakis]{GS}
Bernhard Kerbl, Georgios Kopanas, Thomas Leimk{\"{u}}hler, and George Drettakis.
\newblock 3d gaussian splatting for real-time radiance field rendering.
\newblock \emph{{ACM} Trans. Graph.}, 42\penalty0 (4):\penalty0 139:1--139:14, 2023.

\bibitem[Kirillov et~al.(2023)Kirillov, Mintun, Ravi, Mao, Rolland, Gustafson, Xiao, Whitehead, Berg, Lo, Doll{\'{a}}r, and Girshick]{SAM}
Alexander Kirillov, Eric Mintun, Nikhila Ravi, Hanzi Mao, Chlo{\'{e}} Rolland, Laura Gustafson, Tete Xiao, Spencer Whitehead, Alexander~C. Berg, Wan{-}Yen Lo, Piotr Doll{\'{a}}r, and Ross~B. Girshick.
\newblock Segment anything.
\newblock In \emph{ICCV}, pages 3992--4003, 2023.

\bibitem[Li et~al.(2023)Li, Zhang, Sun, Zou, Liu, Yang, Li, Zhang, and Gao]{semanticSAM}
Feng Li, Hao Zhang, Peize Sun, Xueyan Zou, Shilong Liu, Jianwei Yang, Chunyuan Li, Lei Zhang, and Jianfeng Gao.
\newblock Semantic-sam: Segment and recognize anything at any granularity.
\newblock \emph{arXiv preprint arXiv:2307.04767}, 2023.

\bibitem[Li et~al.(2024)Li, Zheng, Rupprecht, and Vedaldi]{dragapart(3D)}
Ruining Li, Chuanxia Zheng, Christian Rupprecht, and Andrea Vedaldi.
\newblock Dragapart: Learning a part-level motion prior for articulated objects.
\newblock In \emph{ECCV}, pages 165--183, 2024.

\bibitem[Li et~al.(2022)Li, Lin, Zhou, Qi, Wang, and Jia]{MAT}
Wenbo Li, Zhe Lin, Kun Zhou, Lu Qi, Yi Wang, and Jiaya Jia.
\newblock {MAT:} mask-aware transformer for large hole image inpainting.
\newblock In \emph{CVPR}, pages 10748--10758, 2022.

\bibitem[Liu et~al.(2024{\natexlab{a}})Liu, Wang, Cao, Jia, and Huang]{fpe}
Bingyan Liu, Chengyu Wang, Tingfeng Cao, Kui Jia, and Jun Huang.
\newblock Towards understanding cross and self-attention in stable diffusion for text-guided image editing.
\newblock In \emph{CVPR}, pages 7817--7826. {IEEE}, 2024{\natexlab{a}}.

\bibitem[Liu et~al.(2024{\natexlab{b}})Liu, Xu, Yang, Zeng, and He]{DragNoise}
Haofeng Liu, Chenshu Xu, Yifei Yang, Lihua Zeng, and Shengfeng He.
\newblock Drag your noise: Interactive point-based editing via diffusion semantic propagation.
\newblock In \emph{CVPR}, pages 6743--6752, 2024{\natexlab{b}}.

\bibitem[Liu et~al.(2023{\natexlab{a}})Liu, Wu, Hoorick, Tokmakov, Zakharov, and Vondrick]{zero123}
Ruoshi Liu, Rundi Wu, Basile~Van Hoorick, Pavel Tokmakov, Sergey Zakharov, and Carl Vondrick.
\newblock Zero-1-to-3: Zero-shot one image to 3d object.
\newblock In \emph{ICCV}, pages 9264--9275. {IEEE}, 2023{\natexlab{a}}.

\bibitem[Liu et~al.(2023{\natexlab{b}})Liu, Gong, and Liu]{rectifiedflow}
Xingchao Liu, Chengyue Gong, and Qiang Liu.
\newblock Flow straight and fast: Learning to generate and transfer data with rectified flow.
\newblock In \emph{ICLR}, 2023{\natexlab{b}}.

\bibitem[Lowe(2004)]{SiFT}
G Lowe.
\newblock Sift-the scale invariant feature transform.
\newblock \emph{Int. j}, 2004.

\bibitem[Lu et~al.(2024)Lu, Li, and Han]{RegionDrag}
Jingyi Lu, Xinghui Li, and Kai Han.
\newblock Regiondrag: Fast region-based image editing with diffusion models.
\newblock In \emph{ECCV}, pages 231--246, 2024.

\bibitem[Lugmayr et~al.(2022)Lugmayr, Danelljan, Romero, Yu, Timofte, and Gool]{repaint}
Andreas Lugmayr, Martin Danelljan, Andr{\'{e}}s Romero, Fisher Yu, Radu Timofte, and Luc~Van Gool.
\newblock Repaint: Inpainting using denoising diffusion probabilistic models.
\newblock In \emph{CVPR}, pages 11451--11461, 2022.

\bibitem[Luo et~al.(2024)Luo, Darrell, Wang, Goldman, and Holynski]{readoutguidance}
Grace Luo, Trevor Darrell, Oliver Wang, Dan~B. Goldman, and Aleksander Holynski.
\newblock Readout guidance: Learning control from diffusion features.
\newblock In \emph{CVPR}, pages 8217--8227, 2024.

\bibitem[Lv et~al.(2024)Lv, Zhang, Wang, Zheng, Zhong, Li, and Nie]{FourierPrior}
Xiaoqian Lv, Shengping Zhang, Chenyang Wang, Yichen Zheng, Bineng Zhong, Chongyi Li, and Liqiang Nie.
\newblock Fourier priors-guided diffusion for zero-shot joint low-light enhancement and deblurring.
\newblock In \emph{CVPR}, pages 25378--25388, 2024.

\bibitem[Ma et~al.(2017)Ma, Jia, Sun, Schiele, Tuytelaars, and Gool]{PoseGuide}
Liqian Ma, Xu Jia, Qianru Sun, Bernt Schiele, Tinne Tuytelaars, and Luc~Van Gool.
\newblock Pose guided person image generation.
\newblock In \emph{NeurIPS}, pages 406--416, 2017.

\bibitem[Mildenhall et~al.(2020)Mildenhall, Srinivasan, Tancik, Barron, Ramamoorthi, and Ng]{Nerf}
Ben Mildenhall, Pratul~P. Srinivasan, Matthew Tancik, Jonathan~T. Barron, Ravi Ramamoorthi, and Ren Ng.
\newblock Nerf: Representing scenes as neural radiance fields for view synthesis.
\newblock In \emph{ECCV}, pages 405--421. Springer, 2020.

\bibitem[Mou et~al.(2024)Mou, Wang, Song, Shan, and Zhang]{Dragon}
Chong Mou, Xintao Wang, Jiechong Song, Ying Shan, and Jian Zhang.
\newblock Dragondiffusion: Enabling drag-style manipulation on diffusion models.
\newblock In \emph{ICLR}, 2024.

\bibitem[Mu et~al.(2024)Mu, Gharbi, Zhang, Shechtman, Vasconcelos, Wang, and Park]{ImageElements}
Jiteng Mu, Micha{\"{e}}l Gharbi, Richard Zhang, Eli Shechtman, Nuno Vasconcelos, Xiaolong Wang, and Taesung Park.
\newblock Editable image elements for controllable synthesis.
\newblock In \emph{ECCV}, pages 39--56, 2024.

\bibitem[Pan et~al.(2023)Pan, Tewari, Leimk{\"{u}}hler, Liu, Meka, and Theobalt]{DragGAN}
Xingang Pan, Ayush Tewari, Thomas Leimk{\"{u}}hler, Lingjie Liu, Abhimitra Meka, and Christian Theobalt.
\newblock Drag your {GAN:} interactive point-based manipulation on the generative image manifold.
\newblock In \emph{ACM SIGGRAPH}, pages 78:1--78:11, 2023.

\bibitem[Pandey et~al.(2024)Pandey, Guerrero, Gadelha, Hold{-}Geoffroy, Singh, and Mitra]{Diffusionhandles}
Karran Pandey, Paul Guerrero, Matheus Gadelha, Yannick Hold{-}Geoffroy, Karan Singh, and Niloy~J. Mitra.
\newblock Diffusion handles enabling 3d edits for diffusion models by lifting activations to 3d.
\newblock In \emph{CVPR}, pages 7695--7704. {IEEE}, 2024.

\bibitem[Podell et~al.(2024)Podell, English, Lacey, Blattmann, Dockhorn, M{\"{u}}ller, Penna, and Rombach]{SDXL}
Dustin Podell, Zion English, Kyle Lacey, Andreas Blattmann, Tim Dockhorn, Jonas M{\"{u}}ller, Joe Penna, and Robin Rombach.
\newblock {SDXL:} improving latent diffusion models for high-resolution image synthesis.
\newblock In \emph{ICLR}, 2024.

\bibitem[Radford et~al.(2021)Radford, Kim, Hallacy, Ramesh, Goh, Agarwal, Sastry, Askell, Mishkin, Clark, Krueger, and Sutskever]{CLIP}
Alec Radford, Jong~Wook Kim, Chris Hallacy, Aditya Ramesh, Gabriel Goh, Sandhini Agarwal, Girish Sastry, Amanda Askell, Pamela Mishkin, Jack Clark, Gretchen Krueger, and Ilya Sutskever.
\newblock Learning transferable visual models from natural language supervision.
\newblock In \emph{ICML}, pages 8748--8763, 2021.

\bibitem[Ramesh et~al.(2021)Ramesh, Pavlov, Goh, Gray, Voss, Radford, Chen, and Sutskever]{dalle1}
Aditya Ramesh, Mikhail Pavlov, Gabriel Goh, Scott Gray, Chelsea Voss, Alec Radford, Mark Chen, and Ilya Sutskever.
\newblock Zero-shot text-to-image generation.
\newblock In \emph{ICML}, pages 8821--8831, 2021.

\bibitem[Ramesh et~al.(2022)Ramesh, Dhariwal, Nichol, Chu, and Chen]{dalle2}
Aditya Ramesh, Prafulla Dhariwal, Alex Nichol, Casey Chu, and Mark Chen.
\newblock Hierarchical text-conditional image generation with {CLIP} latents.
\newblock \emph{CoRR}, abs/2204.06125, 2022.

\bibitem[Ren et~al.(2024{\natexlab{a}})Ren, Xu, Wu, Liu, Xiang, and Toisoul]{SceneDiffusion}
Jiawei Ren, Mengmeng Xu, Jui{-}Chieh Wu, Ziwei Liu, Tao Xiang, and Antoine Toisoul.
\newblock Move anything with layered scene diffusion.
\newblock In \emph{CVPR}, pages 6380--6389, 2024{\natexlab{a}}.

\bibitem[Ren et~al.(2024{\natexlab{b}})Ren, Liu, Zeng, Lin, Li, Cao, Chen, Huang, Chen, Yan, Zeng, Zhang, Li, Yang, Li, Jiang, and Zhang]{GroundingSAM}
Tianhe Ren, Shilong Liu, Ailing Zeng, Jing Lin, Kunchang Li, He Cao, Jiayu Chen, Xinyu Huang, Yukang Chen, Feng Yan, Zhaoyang Zeng, Hao Zhang, Feng Li, Jie Yang, Hongyang Li, Qing Jiang, and Lei Zhang.
\newblock Grounded {SAM:} assembling open-world models for diverse visual tasks.
\newblock \emph{arXiv preprint arXiv:2401.14159}, abs/2401.14159, 2024{\natexlab{b}}.

\bibitem[Rombach et~al.(2022)Rombach, Blattmann, Lorenz, Esser, and Ommer]{SD}
Robin Rombach, Andreas Blattmann, Dominik Lorenz, Patrick Esser, and Bj{\"{o}}rn Ommer.
\newblock High-resolution image synthesis with latent diffusion models.
\newblock In \emph{CVPR}, pages 10674--10685, 2022.

\bibitem[Saharia et~al.(2022)Saharia, Chan, Saxena, Li, Whang, Denton, Ghasemipour, Lopes, Ayan, Salimans, Ho, Fleet, and Norouzi]{imagen}
Chitwan Saharia, William Chan, Saurabh Saxena, Lala Li, Jay Whang, Emily~L. Denton, Seyed Kamyar~Seyed Ghasemipour, Raphael~Gontijo Lopes, Burcu~Karagol Ayan, Tim Salimans, Jonathan Ho, David~J. Fleet, and Mohammad Norouzi.
\newblock Photorealistic text-to-image diffusion models with deep language understanding.
\newblock In \emph{NeurIPS}, 2022.

\bibitem[Sajnani et~al.(2025)Sajnani, van Baar, Min, Katyal, and Sridhar]{geodiffuser}
Rahul Sajnani, Jeroen van Baar, Jie Min, Kapil Katyal, and Srinath Sridhar.
\newblock Geodiffuser: Geometry-based image editing with diffusion models.
\newblock In \emph{WACV}, pages 472--482, 2025.

\bibitem[Seitzer(2020)]{pytorch-FID}
Maximilian Seitzer.
\newblock {pytorch-fid: FID Score for PyTorch}.
\newblock \url{https://github.com/mseitzer/pytorch-fid}, 2020.
\newblock Version 0.3.0.

\bibitem[Shi et~al.(2024)Shi, Xue, Liew, Pan, Yan, Zhang, Tan, and Bai]{dragdiffusion}
Yujun Shi, Chuhui Xue, Jun~Hao Liew, Jiachun Pan, Hanshu Yan, Wenqing Zhang, Vincent Y.~F. Tan, and Song Bai.
\newblock Dragdiffusion: Harnessing diffusion models for interactive point-based image editing.
\newblock In \emph{CVPR}, 2024.

\bibitem[Shin et~al.(2024)Shin, Choi, and Park]{InstantDrag}
Joonghyuk Shin, Daehyeon Choi, and Jaesik Park.
\newblock Instantdrag: Improving interactivity in drag-based image editing.
\newblock In \emph{ACM SIGGRAPH}, pages 39:1--39:10, 2024.

\bibitem[Siarohin et~al.(2018)Siarohin, Sangineto, Lathuili{\`{e}}re, and Sebe]{DefromGAN}
Aliaksandr Siarohin, Enver Sangineto, St{\'{e}}phane Lathuili{\`{e}}re, and Nicu Sebe.
\newblock Deformable gans for pose-based human image generation.
\newblock In \emph{CVPR}, pages 3408--3416. Computer Vision Foundation / {IEEE} Computer Society, 2018.

\bibitem[Song et~al.(2021)Song, Meng, and Ermon]{DDIM}
Jiaming Song, Chenlin Meng, and Stefano Ermon.
\newblock Denoising diffusion implicit models.
\newblock In \emph{ICLR}, 2021.

\bibitem[Stacchio(2023)]{stableinpainting}
Lorenzo Stacchio.
\newblock Train stable diffusion for inpainting, 2023.

\bibitem[Stein et~al.(2023)Stein, Cresswell, Hosseinzadeh, Sui, Ross, Villecroze, Liu, Caterini, Taylor, and Loaiza{-}Ganem]{DINOv2}
George Stein, Jesse~C. Cresswell, Rasa Hosseinzadeh, Yi Sui, Brendan~Leigh Ross, Valentin Villecroze, Zhaoyan Liu, Anthony~L. Caterini, J.~Eric~T. Taylor, and Gabriel Loaiza{-}Ganem.
\newblock Exposing flaws of generative model evaluation metrics and their unfair treatment of diffusion models.
\newblock In \emph{NeurIPS}, 2023.

\bibitem[Sun et~al.(2025)Sun, Dong, Cui, and Tang]{attentiveEraser}
Wenhao Sun, Xue{-}Mei Dong, Benlei Cui, and Jingqun Tang.
\newblock Attentive eraser: Unleashing diffusion model's object removal potential via self-attention redirection guidance.
\newblock In \emph{AAAI}, pages 20734--20742, 2025.

\bibitem[Suvorov et~al.(2022)Suvorov, Logacheva, Mashikhin, Remizova, Ashukha, Silvestrov, Kong, Goka, Park, and Lempitsky]{Lama}
Roman Suvorov, Elizaveta Logacheva, Anton Mashikhin, Anastasia Remizova, Arsenii Ashukha, Aleksei Silvestrov, Naejin Kong, Harshith Goka, Kiwoong Park, and Victor Lempitsky.
\newblock Resolution-robust large mask inpainting with fourier convolutions.
\newblock In \emph{WACV}, pages 3172--3182, 2022.

\bibitem[Szegedy et~al.(2016)Szegedy, Vanhoucke, Ioffe, Shlens, and Wojna]{inception}
Christian Szegedy, Vincent Vanhoucke, Sergey Ioffe, Jon Shlens, and Zbigniew Wojna.
\newblock Rethinking the inception architecture for computer vision.
\newblock In \emph{CVPR}, 2016.

\bibitem[Tan et~al.(2024)Tan, Liu, Yang, Xue, and Wang]{Subjects200k}
Zhenxiong Tan, Songhua Liu, Xingyi Yang, Qiaochu Xue, and Xinchao Wang.
\newblock Ominicontrol: Minimal and universal control for diffusion transformer.
\newblock \emph{CoRR}, abs/2411.15098, 2024.

\bibitem[Tang et~al.(2023)Tang, Jia, Wang, Phoo, and Hariharan]{DiFT}
Luming Tang, Menglin Jia, Qianqian Wang, Cheng~Perng Phoo, and Bharath Hariharan.
\newblock Emergent correspondence from image diffusion.
\newblock In \emph{NeurIPS}, 2023.

\bibitem[Teed and Deng(2020)]{RAFT}
Zachary Teed and Jia Deng.
\newblock {RAFT:} recurrent all-pairs field transforms for optical flow.
\newblock In \emph{ICCV}, pages 402--419, 2020.

\bibitem[Tewel et~al.(2024)Tewel, Kaduri, Gal, Kasten, Wolf, Chechik, and Atzmon]{ConsiStory}
Yoad Tewel, Omri Kaduri, Rinon Gal, Yoni Kasten, Lior Wolf, Gal Chechik, and Yuval Atzmon.
\newblock Training-free consistent text-to-image generation.
\newblock \emph{ACM Transactions on Graphics (TOG)}, 43\penalty0 (4):\penalty0 1--18, 2024.

\bibitem[Voleti et~al.(2024)Voleti, Yao, Boss, Letts, Pankratz, Tochilkin, Laforte, Rombach, and Jampani]{SV3D}
Vikram Voleti, Chun{-}Han Yao, Mark Boss, Adam Letts, David Pankratz, Dmitry Tochilkin, Christian Laforte, Robin Rombach, and Varun Jampani.
\newblock {SV3D:} novel multi-view synthesis and 3d generation from a single image using latent video diffusion.
\newblock In \emph{ECCV}, pages 439--457, 2024.

\bibitem[Wang et~al.(2018)Wang, Zheng, Liang, Chen, Lin, and Yang]{CAP-VITON}
Bochao Wang, Huabin Zheng, Xiaodan Liang, Yimin Chen, Liang Lin, and Meng Yang.
\newblock Toward characteristic-preserving image-based virtual try-on network.
\newblock In \emph{ECCV}, pages 607--623. Springer, 2018.

\bibitem[Wang et~al.(2024)Wang, Mei, and Yuille]{SCLIP}
Feng Wang, Jieru Mei, and Alan~L. Yuille.
\newblock {SCLIP:} rethinking self-attention for dense vision-language inference.
\newblock In \emph{ECCV}, pages 315--332, 2024.

\bibitem[Xiao et~al.(2024)Xiao, Feng, Zhang, Liu, Yang, Zhu, Fu, Zhu, Liu, and Zha]{dreamclean}
Jie Xiao, Ruili Feng, Han Zhang, Zhiheng Liu, Zhantao Yang, Yurui Zhu, Xueyang Fu, Kai Zhu, Yu Liu, and Zheng{-}Jun Zha.
\newblock Dreamclean: Restoring clean image using deep diffusion prior.
\newblock In \emph{ICLR}, 2024.

\bibitem[Yang et~al.(2024{\natexlab{a}})Yang, Kang, Huang, Xu, Feng, and Zhao]{DepthAnything}
Lihe Yang, Bingyi Kang, Zilong Huang, Xiaogang Xu, Jiashi Feng, and Hengshuang Zhao.
\newblock Depth anything: Unleashing the power of large-scale unlabeled data.
\newblock In \emph{CVPR}, pages 10371--10381, 2024{\natexlab{a}}.

\bibitem[Yang et~al.(2024{\natexlab{b}})Yang, Zhang, Song, Hong, Xu, Zhao, Zhang, Cui, and Yang]{diffusionsurvey}
Ling Yang, Zhilong Zhang, Yang Song, Shenda Hong, Runsheng Xu, Yue Zhao, Wentao Zhang, Bin Cui, and Ming{-}Hsuan Yang.
\newblock Diffusion models: {A} comprehensive survey of methods and applications.
\newblock \emph{{ACM} Comput. Surv.}, 56\penalty0 (4):\penalty0 105:1--105:39, 2024{\natexlab{b}}.

\bibitem[Yenphraphai et~al.(2024)Yenphraphai, Pan, Liu, Panozzo, and Xie]{ImageSculpting}
Jiraphon Yenphraphai, Xichen Pan, Sainan Liu, Daniele Panozzo, and Saining Xie.
\newblock Image sculpting: Precise object editing with 3d geometry control.
\newblock In \emph{CVPR}, pages 4241--4251, 2024.

\bibitem[Zhou()]{self-guide-implement}
Shengzhe Zhou.
\newblock Diffusion self guidance implementation.

\bibitem[Zhu et~al.(2023)Zhu, Zhang, Liang, Cao, Wen, Timofte, and Gool]{prior_restoration}
Yuanzhi Zhu, Kai Zhang, Jingyun Liang, Jiezhang Cao, Bihan Wen, Radu Timofte, and Luc~Van Gool.
\newblock Denoising diffusion models for plug-and-play image restoration.
\newblock In \emph{CVPR}, pages 1219--1229, 2023.

\bibitem[Zhu et~al.(2019)Zhu, Huang, Shi, Yu, Wang, and Bai]{ProgressivePoseTransfer}
Zhen Zhu, Tengteng Huang, Baoguang Shi, Miao Yu, Bofei Wang, and Xiang Bai.
\newblock Progressive pose attention transfer for person image generation.
\newblock In \emph{CVPR}, pages 2347--2356, 2019.

\bibitem[Zhu et~al.(2022)Zhu, Huang, Xu, Shi, Cheng, and Bai]{ProgressiveAlignPoseTransfer}
Zhen Zhu, Tengteng Huang, Mengde Xu, Baoguang Shi, Wenqing Cheng, and Xiang Bai.
\newblock Progressive and aligned pose attention transfer for person image generation.
\newblock \emph{{IEEE} Trans. Pattern Anal. Mach. Intell.}, 44\penalty0 (8):\penalty0 4306--4320, 2022.

\end{thebibliography}
}


\clearpage
\setcounter{page}{1}
\maketitlesupplementary

\appendix

\section{Additional Results}\label{sec:more_results}
\begin{figure*}[p]
  \centering
  \includegraphics[width=0.95\textwidth]{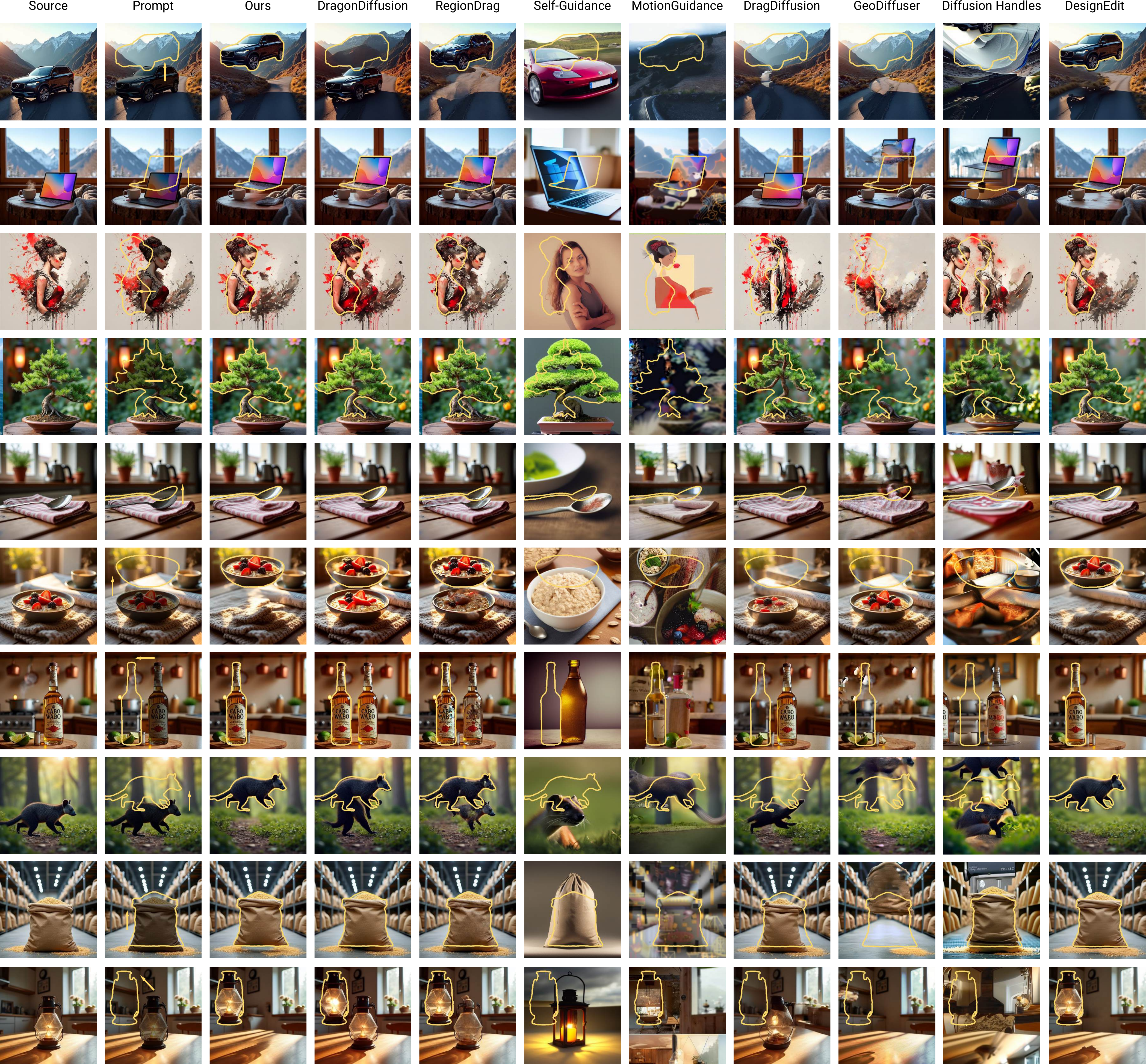}
  \caption{Qualitative comparison with state-of-the-art editing methods in moving operations.}
  \label{fig_sup:move}
\end{figure*}

\begin{figure*}[p]
  \centering
  \includegraphics[width=0.95\textwidth]{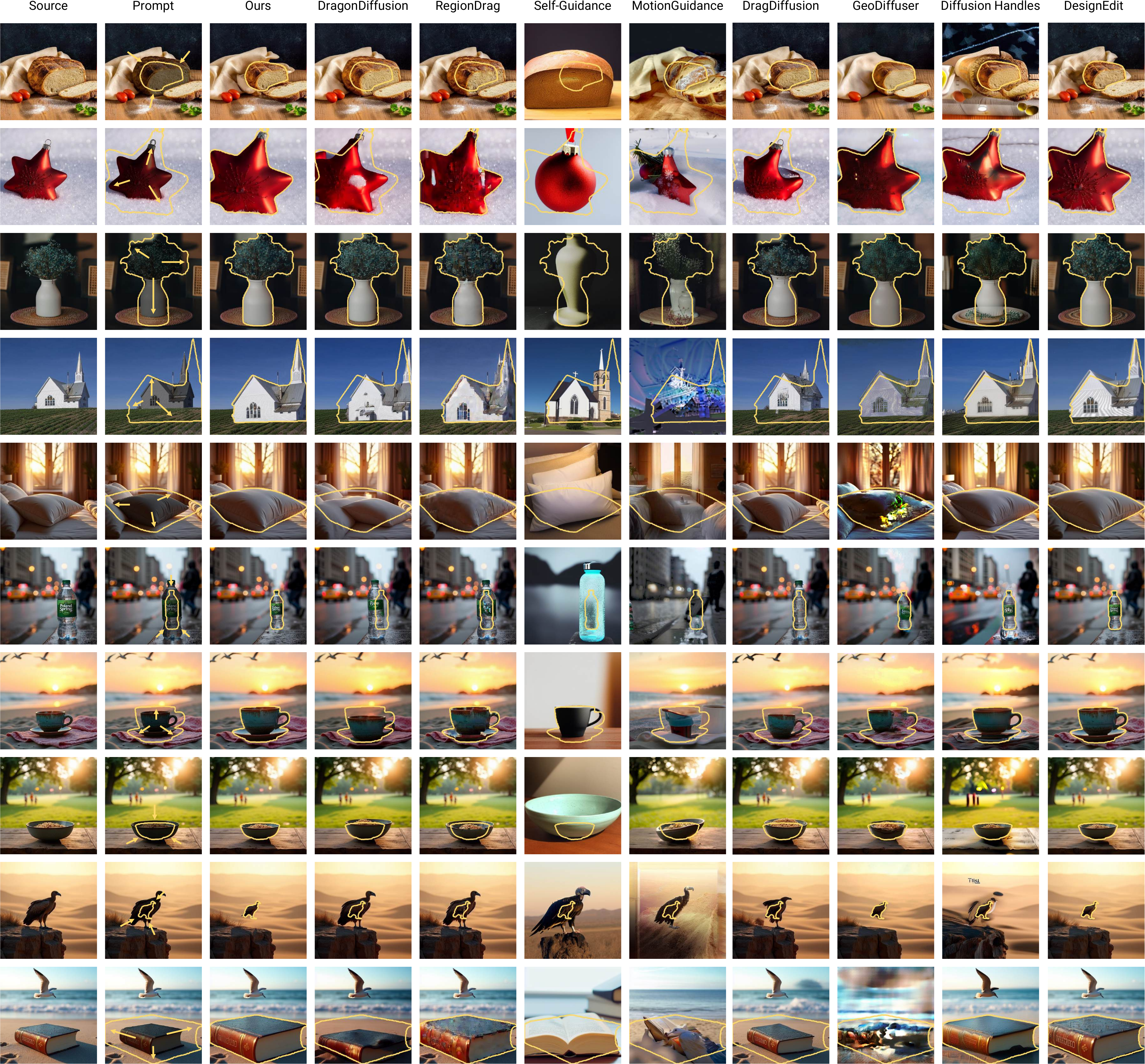}
  \caption{Qualitative comparison with state-of-the-art editing methods in scaling operation.}
  \label{fig_sup:resize}
\end{figure*}

\begin{figure*}[p]
  \centering
  \includegraphics[width=0.95\textwidth]{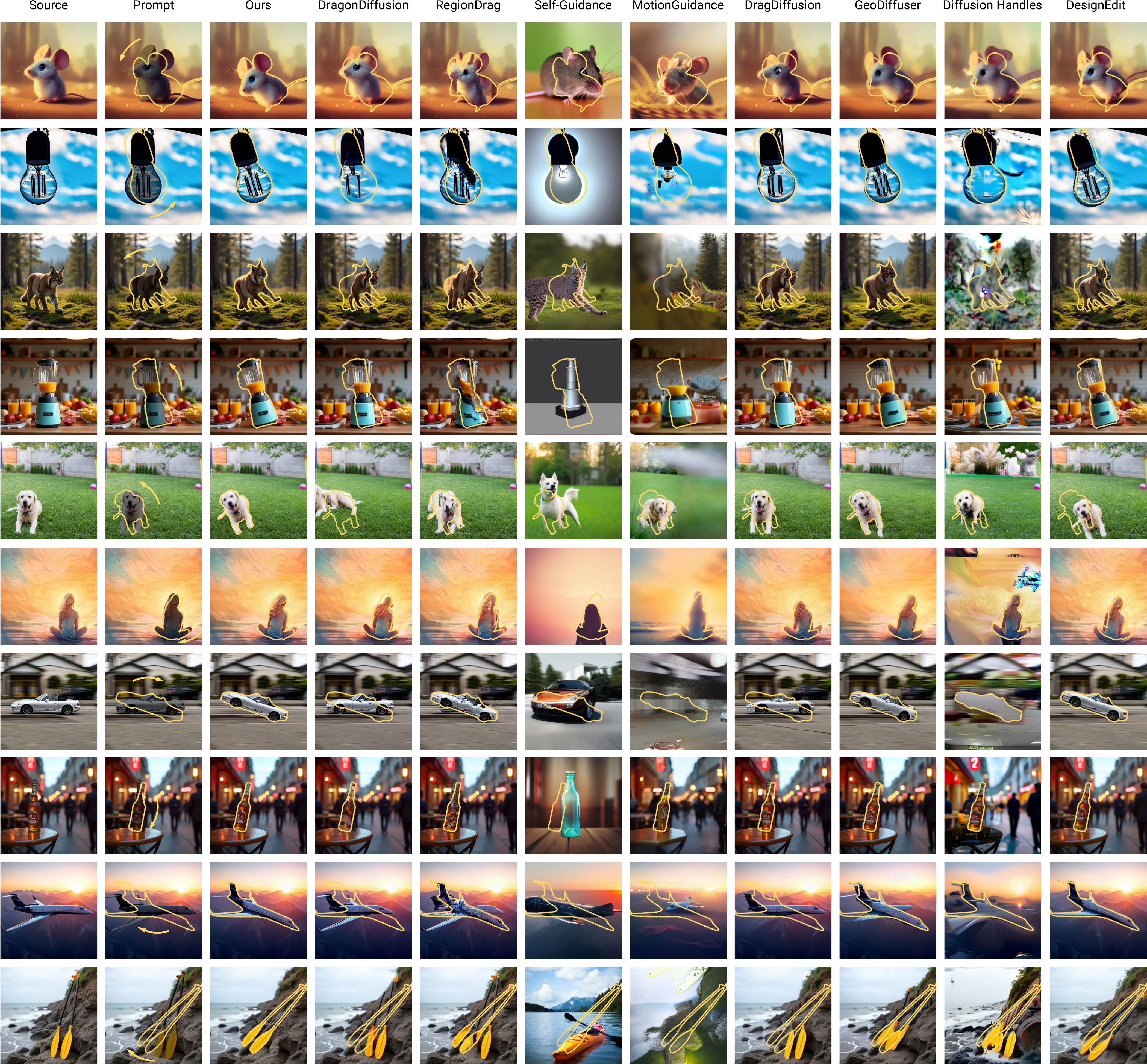}
  \caption{Qualitative comparison with state-of-the-art editing methods in rotation operation.}
  \label{fig_sup:rotate}
\end{figure*}

\begin{figure*}[ht]
  \centering
  \includegraphics[width=0.85\textwidth]{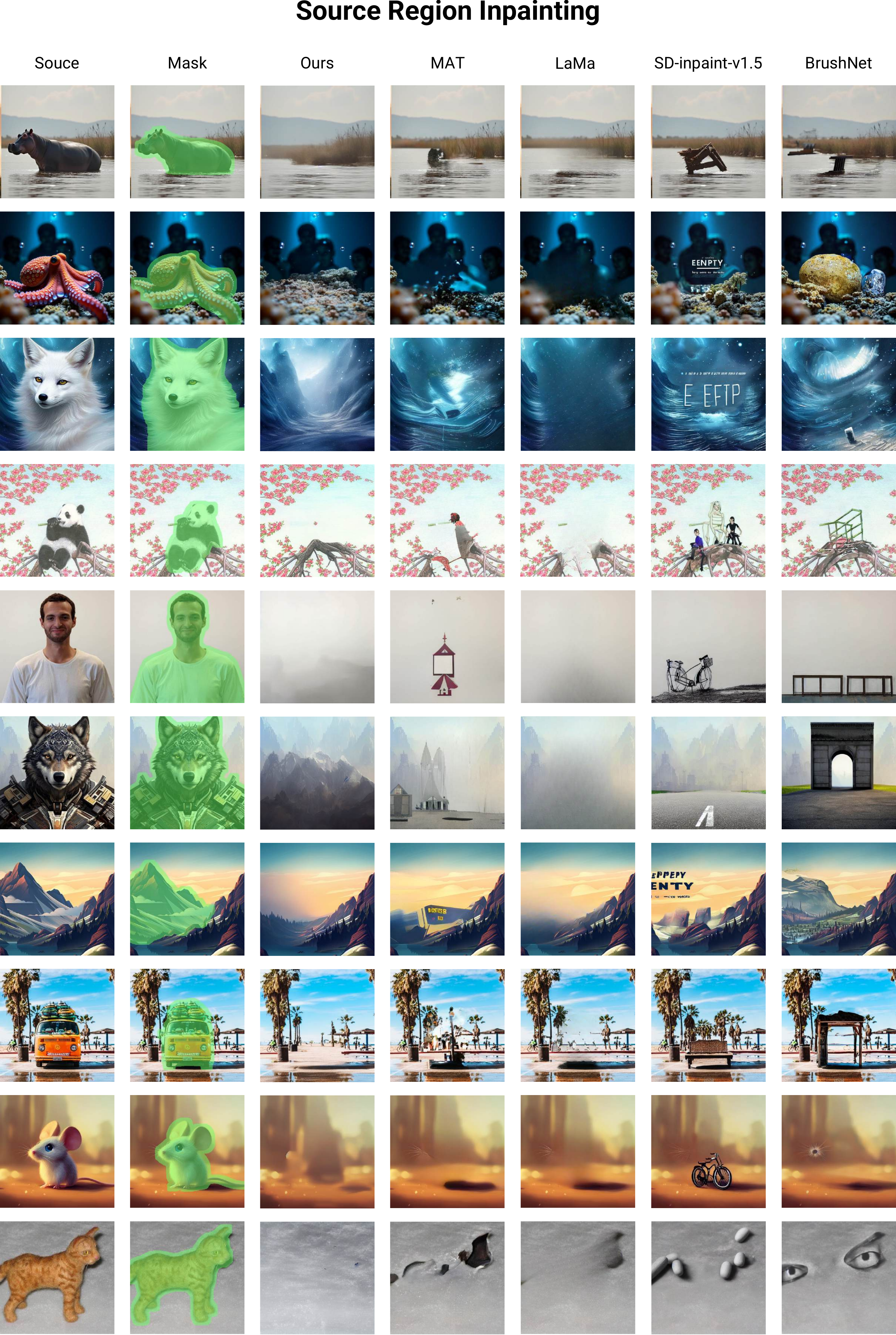}
  \caption{Qualitative comparison with state-of-the-art inpainting methods in source region inpainting.}
  \label{fig_sup:inpaint_src}
\end{figure*}

\begin{figure*}[ht]
  \centering
  \includegraphics[width=0.8\textwidth]{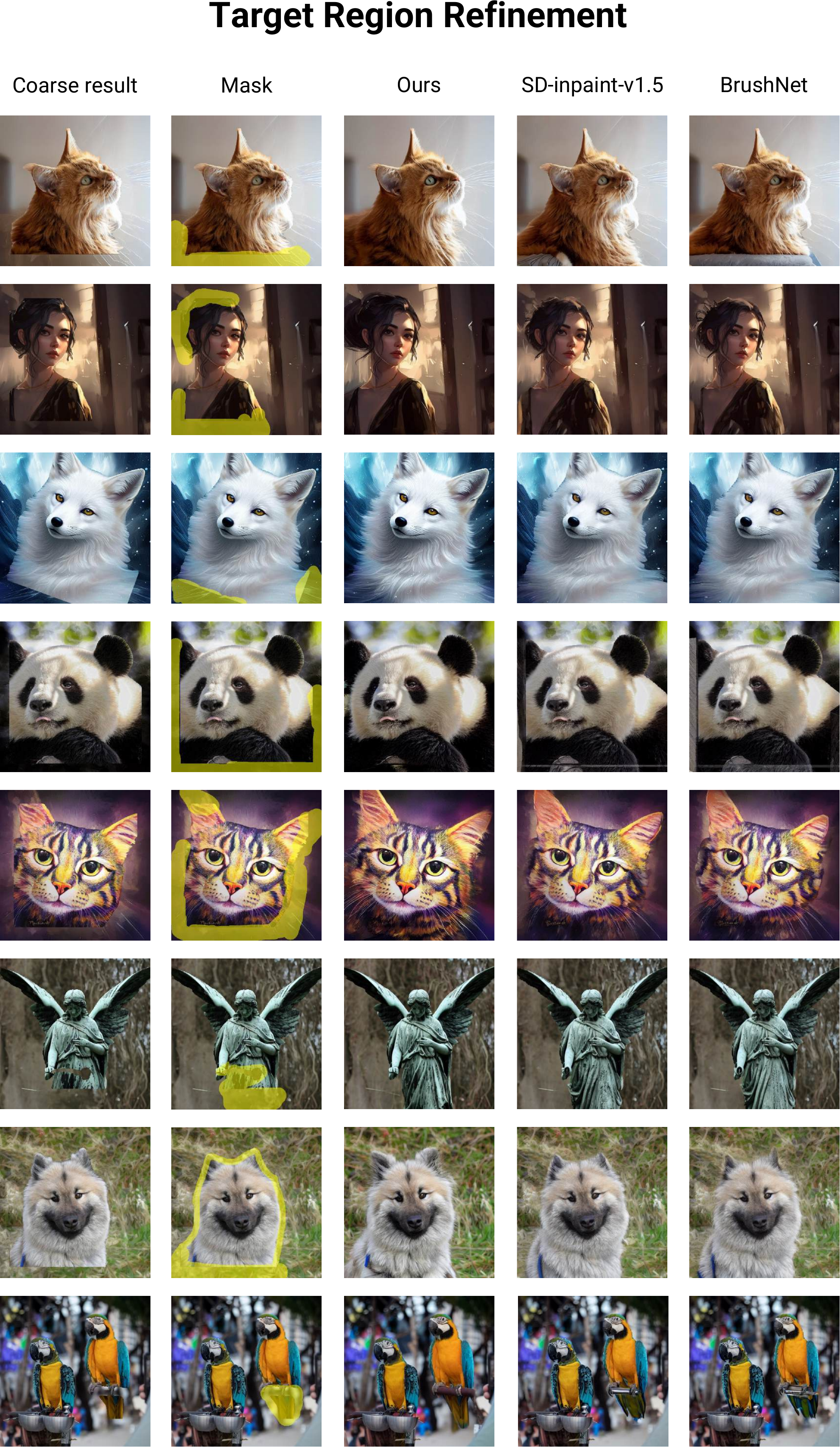}
  \caption{Qualitative comparison with state-of-the-art inpainting methods in target region refinement.}
  \label{fig_sup:inpaint_tgt}
\end{figure*}

We provide additional qualitative comparisons to state-of-the-art methods in Fig.~\ref{fig_sup:move}, Fig.~\ref{fig_sup:resize}, and Fig.~\ref{fig_sup:rotate} for 2D editing tasks, Fig.~\ref{fig_sup:3d} for 3D editing tasks, and Fig.~\ref{fig_sup:inpaint_src}, Fig.~\ref{fig_sup:inpaint_tgt} for inpainting tasks.

\section{Extended Applications}
\begin{figure}[ht]
  \centering
  \includegraphics[width=0.45\textwidth]{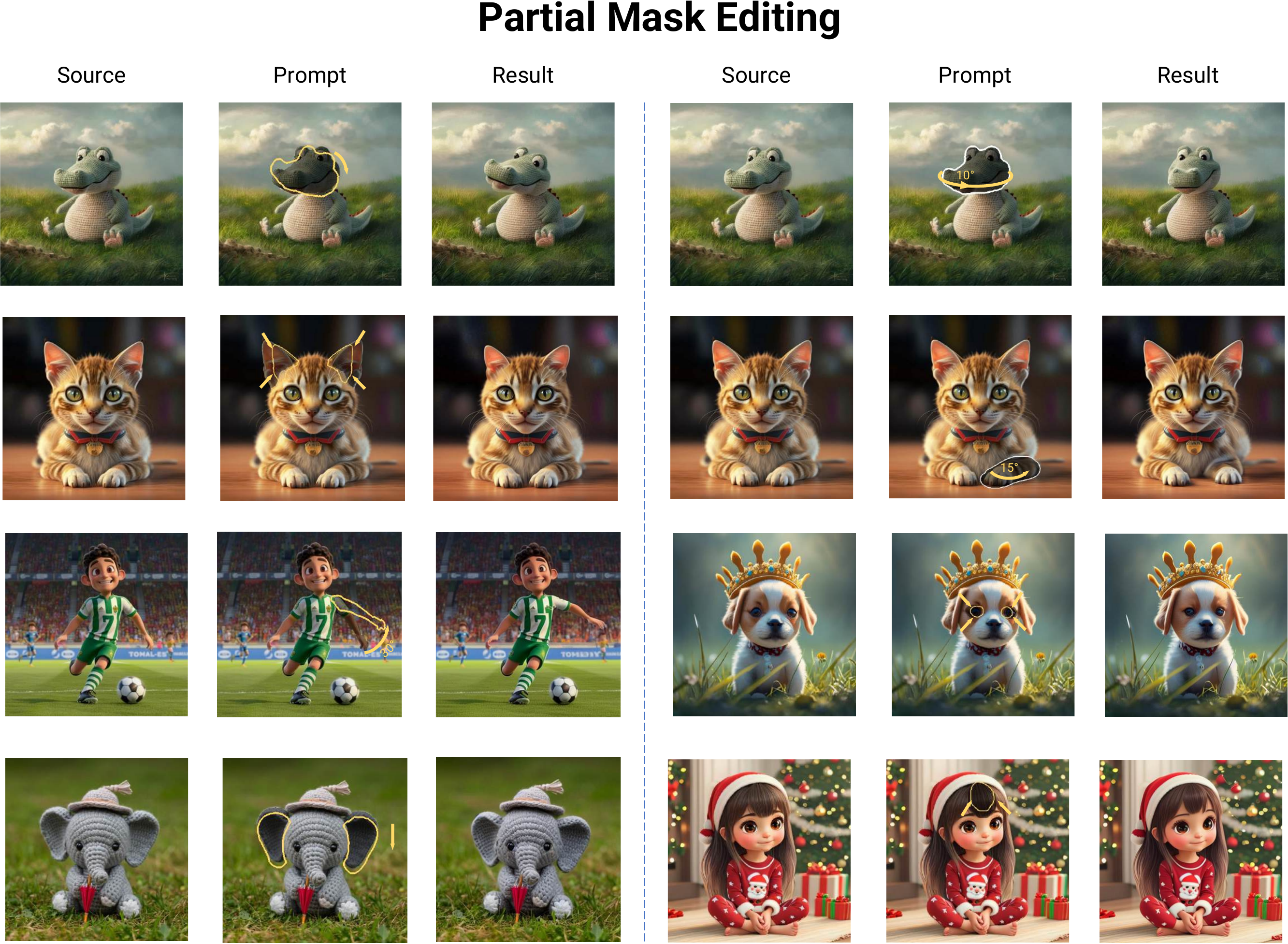}
  \caption{Qualitative result on Partial Mask editing tasks. Instead of editing a whole object, this task focuses on editing parts of an object.}
  \label{fig_sup:partial}
\end{figure}

\begin {figure*}[ht]
\centering
\begin {minipage}[t]{0.45\textwidth}
\centering
\includegraphics [width=\textwidth]{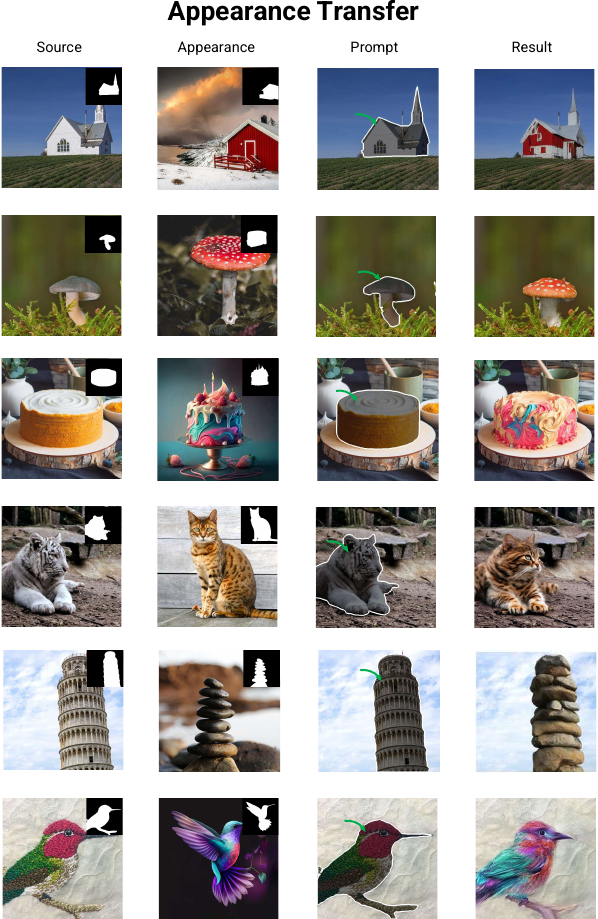}
\end {minipage}
\hspace {5mm} 
\begin {minipage}[t]{0.45\textwidth}
\centering
\includegraphics [width=\textwidth]{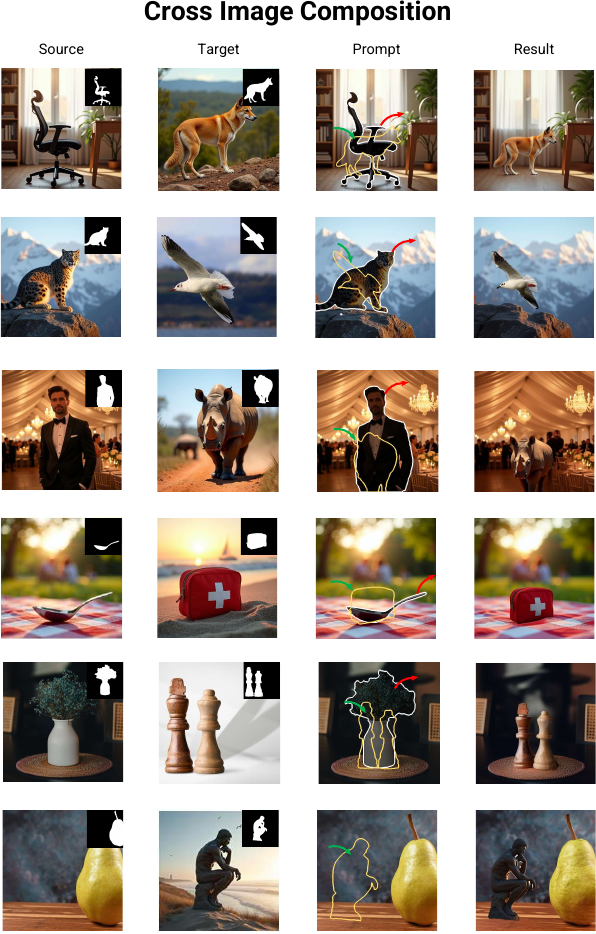}
\end {minipage}
\caption {Qualitative results on Extended Applications}
\label {fig_sup:merge}
\end {figure*}

Beyond the core tasks presented in the main paper, \pipeline{} further supports \textbf{Partial Mask Editing}, \textbf{Appearance Transfer}, and \textbf{Cross-Image Composition} to address more complex editing scenarios. Qualitative results for these extended capabilities are shown in Fig.~\ref{fig_sup:partial} and Fig.~\ref{fig_sup:merge} , collectively demonstrating the versatility of FreeFine. We formalize the key implementation details as follows:

\noindent{\bf{Appearance Transfer}} aims to preserve the shape of the source object while replacing its textural or color properties with those of a target object. This is achieved using our TCA module: specifically, by substituting the source image $I_s$ and source mask $M_s$ with the target appearance image and its corresponding object mask, while retaining $M_t$ (the mask of the object to be edited). Additionally, we employ the LP module on $M_t$ to facilitate appearance modification, and can optionally specify the new object category via our CG module to further enhance details.

\noindent{\bf{Cross-Image Composition}} is a generalized geometric editing task that involves rearranging multiple objects within a shared canvas. This is implemented through our editing pipeline in three key steps: (1) Removing original objects from the source image to produce a clean canvas—this step is optional and only performed when existing objects need to be relocated or replaced; (2) Copying target objects into the canvas and adjusting their geometric properties (e.g., repositioning, reorienting, rescaling) to generate a coarse composite image $\hat{I}_c$; (3) Refining the target regions using the same refinement process described in the main paper. 

For composing $N$ objects, the required input parameters are formalized as a set $\mathcal{P}$:
\begin{equation*}
\mathcal{P} = \left\{ \left( I_{s,i}, M_{s,i}, M_{t,i}; \, L_i^*, \, M_{d,i}^* \right) \mid \forall i \in \{1, 2, ..., N\} \right\}
\end{equation*}
where $I_{s,i}$, $M_{s,i}$, and $M_{t,i}$ are mandatory parameters: $I_{s,i}$ denotes the source image of the $i$-th object, $M_{s,i}$ its source mask, and $M_{t,i}$ its target mask in the canvas; superscript $^*$ indicates optional parameters: $L_i^*$ corresponds to the category label for the CG module, and $M_{d,i}^*$ denotes the draw mask of the $i$-th object for structure completion.


\section{Additional Ablation Studies} \label{supp:sec:ablations}

\begin{figure}[h!]
  \centering
  \includegraphics[width=\linewidth]{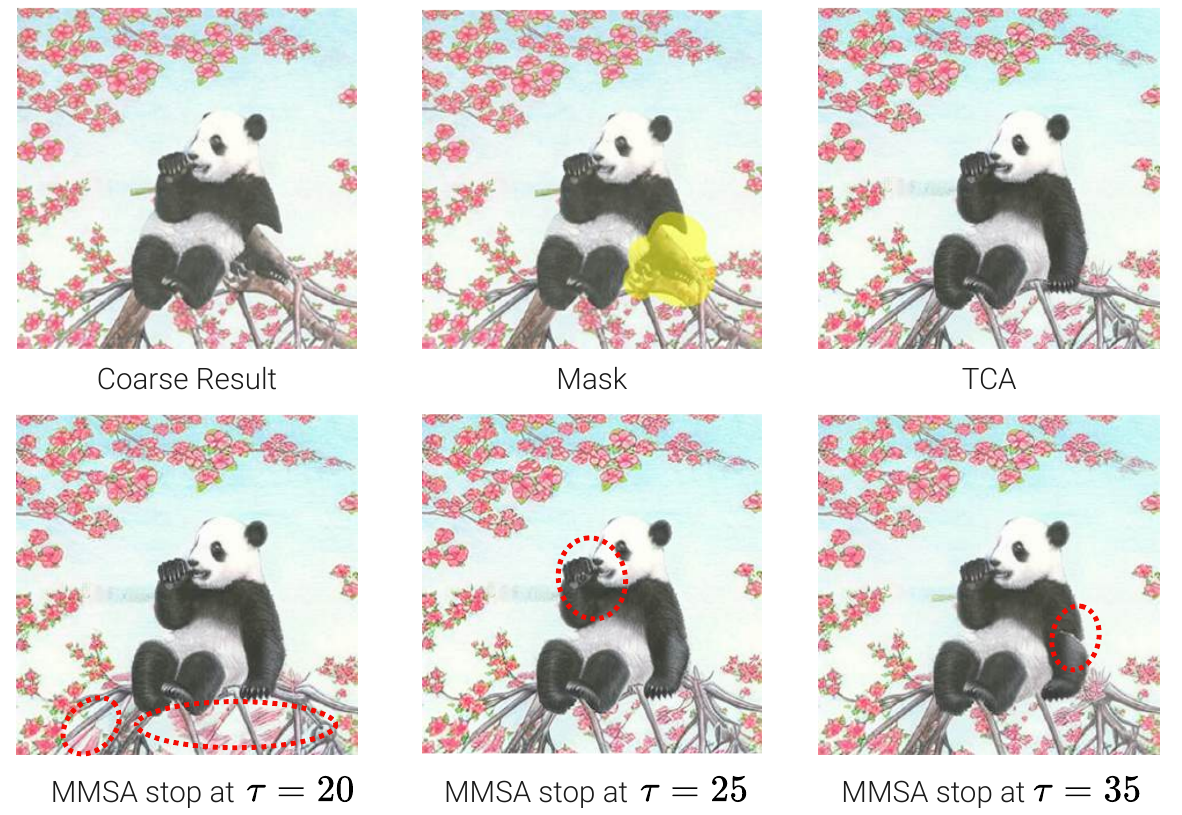}
  \caption{Comparisons between the proposed TCA and the straightforward early stop strategy with MMSA. Experiments are conducted while keeping LP and CG applied at the same scale. }
  \label{fig_sup:early_stop}
\end{figure}

We further compare \attention{} with the timestep thresholding alternative mentioned in the main paper, which is equivalent to early stopping using MMSA after reaching a timestep threshold. As shown in Fig.~\ref{fig_sup:early_stop}, the early-stop strategy faces a trade-off: stopping too early introduces undesired changes, while stopping too late degrades structural completion. In contrast, \attention{} avoids this trade-off by smoothly transitioning between MMSA and self-attention, achieving both better completion and preservation of details.

\section{Additional Implementation Details} \label{supp:sec:impl_details}

\subsection{Step 1: Object Transformation Details}
\label{supp:sec:transform_details}

As described in the main paper, Step~1 of our pipeline involves transforming the object in the source image $I_s$ based on user instructions. Here, we provide additional details, organized from simple 2D edits to complex 3D transformations.

\vspace{1mm}\noindent\textbf{2D Transformations.}~For purely 2D edits, $\mathcal{T}_{\theta}$ represents an affine transformation, which can be expressed as a unified transformation matrix:
\[
\begin{pmatrix}
x' \\
y' \\
1
\end{pmatrix}
=
\begin{pmatrix}
s_x \cdot \cos\phi & -s_y \cdot \sin\phi & t_x \\
s_x \cdot \sin\phi & s_y \cdot \cos\phi & t_y \\
0 & 0 & 1
\end{pmatrix}
\begin{pmatrix}
x \\
y \\
1
\end{pmatrix},
\]
where $(x, y)$ are pixel coordinates in $M_s$, and $(x', y')$ are the transformed coordinates. Here, $s_x$ and $s_y$ represent scaling factors along the $x$ and $y$ axes, $\phi$ is the rotation angle, and $(t_x, t_y)$ is the translation vector. This matrix supports combined operations such as scaling, rotation, and translation in a single transformation.

\vspace{1mm}\noindent\textbf{3D Transformations.}~For 3D edits, we explore two approaches: (1) depth estimation from a single image to establish a sparse 3D representation, lift the object to 3D space, apply transformations, and re-project it back to 2D; and (2) leveraging video models, particularly multi-view video generation from a single image~\cite{SV3D}. Note that existing multi-view generation methods typically require clean foreground images with white backgrounds, and the transformed objects often suffer from detail loss and structural artifacts(e.g. incomplete or missing parts). These limitations highlight the importance of our Step~3 (target region refinement) for achieving high-quality results.

In the first approach, we estimate the scene depth using a depth estimator~\cite{DepthAnything}. Given the depth map $D_s$ of $I_s$, we lift the object in $M_s$ to 3D space using the camera intrinsic matrix $K$:
\[
P_s = K^{-1} \cdot (x, y, 1)^\top \cdot D_s(x, y),
\]
where $P_s$ represents the 3D coordinates of the object. We then apply the transformation $\mathcal{T}_{\theta}$ in 3D space. For example, a rotation along the $y$-axis by $\phi$ degrees is represented as:
\[
P_t = R_y(\phi) \cdot P_s,
\]
where $R_y(\phi)$ is the rotation matrix. Finally, we re-project the 3D points back to 2D image space:
\[
(x', y') = K \cdot P_t.
\]
However, this method has several limitations: (1) the camera intrinsic matrix $K$ is unknown and must be estimated or assumed, (2) the depth map from~\cite{DepthAnything} provides only relative depth, and (3) the sparse 3D representation leads to artifacts during reprojection, requiring additional rasterization and refinement steps. Due to these limitations and its reliance on strong assumptions, this approach struggles to handle large-angle 3D transformations effectively. 

To address the challenge of large-angle 3D transformations, we utilize video models such as~\cite{SV3D} to generate a multi-view video of the target object. Following the approach of~\cite{SV3D}, we specify the elevation angle and generate an $n$-frame multi-view video by controlling the azimuth angles. By indexing the frames, we extract the corresponding transformed images, which are rendered on a white background. We then align the center of the transformed object with the source mask and composite it back into the original image. The key parameters include the azimuth angles (e.g., $[0^\circ, 30^\circ, \dots, 330^\circ]$) and elevation angles (e.g., $[-10^\circ, 0^\circ, 10^\circ]$), which allow us to control the viewpoint and generate diverse transformations

\begin{figure*}[ht]
  \centering
  \includegraphics[width=0.8\textwidth]{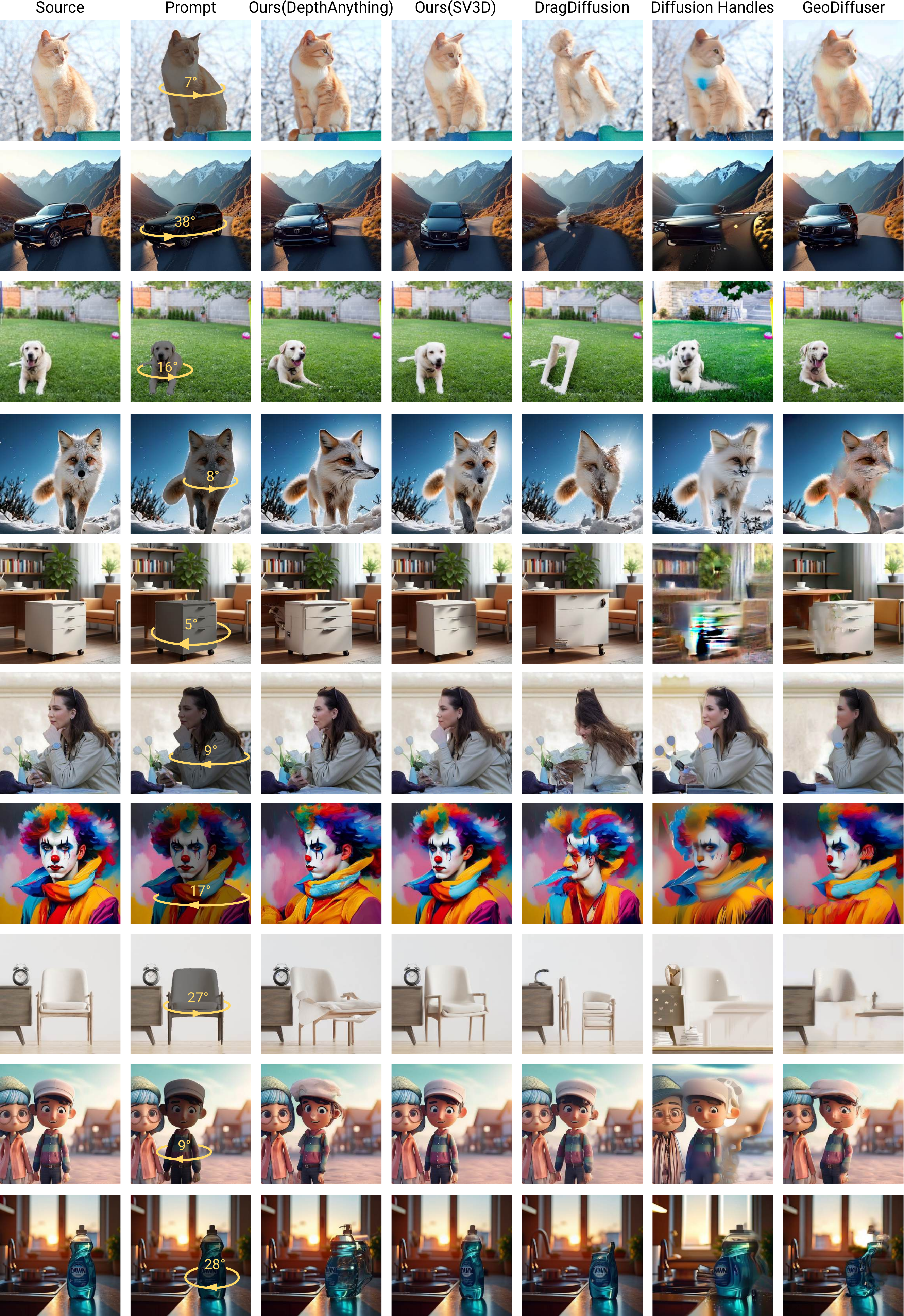}
    \caption{Qualitative comparison with state-of-the-art inpainting methods in 3D-editing scenarios. Two variants of our method are distinguished by suffixes, indicating the different underlying lifting models.}
  \label{fig_sup:3d}
\end{figure*}

Notably, both 3D rotation via video models (e.g., SV3D) and that via depth estimation (e.g., DepthAnything) suffer from imprecise angle control. In the main paper, we focus on fair quantitative comparisons under the same depth-based setup. As shown in Fig.~\ref{fig_sup:3d}, even with identical input angles, results from our SV3D-based approach and depth-based approach show deviations, yet the former yields more realistic editing outcomes.

\vspace{1mm}\noindent\textbf{Flexible Combination.}~Our pipeline supports arbitrary combinations of 2D and 3D transformations, allowing users to achieve both precise control and realistic results. For example, a 2D translation can be combined with a 3D rotation to create complex editing effects.

\subsection{Local Perturbation}\label{supp:sec:lp_details}
As described in the main paper, Local Perturbation (LP) introduces controlled stochasticity within a user-defined mask $\mathcal{M}$ to enable dramatic changes in editing scenarios. Here, we provide additional implementation details.

The original sampling process can be formulated as:
\begin{align}
\hat{x}_0 &= \frac{x_t - \sqrt{1 - \alpha_t} \cdot \epsilon_{\theta}(x_t, t)}{\sqrt{\alpha_t}}, \\
\tilde{x}_{t-1} &= \sqrt{\alpha_{t-1}} \cdot \tilde{x}_0 + \sqrt{1 - \alpha_{t-1} - \sigma_t^2} \cdot \epsilon_{\theta}(x_t, t) + \sigma_t \cdot \epsilon,
\end{align}
where \( t \) is the timestep, \( x_t \) is the latent variable, \( \epsilon_{\theta}(x_t, t) \) is the model-predicted noise, \( \alpha_t \) is a noise scheduling parameter, \( \epsilon \sim \mathcal{N}(0, I) \) is standard Gaussian noise, and \( \sigma_t \) is set to \( 0 \) for deterministic control.

Our key insight is to introduce additional stochasticity into local completion regions while maintaining deterministic behavior elsewhere. Specifically, we set:
\begin{align}
  \sigma_t &= 
  \begin{cases} 
    \sqrt{\frac{1 - \alpha_{t-1}}{1 - \alpha_t} \cdot \left( 1 - \frac{\alpha_t}{\alpha_{t-1}} \right)}, & \text{if } \mathcal{M} \\
    0, & \text{otherwise}.
  \end{cases}
\end{align}
This allows LP to selectively apply DDPM updates~\cite{DDPM} within the mask and DDIM updates~\cite{DDIM} elsewhere, balancing flexibility and control.

LP is versatile and can be applied to various tasks with different purposes. In Step~2 (source region inpainting), LP is applied to the source mask $M_s$ to remove unwanted objects while preserving the background. In Step~3 (target region refinement), LP is applied to the target mask $M_d$ to refine and complete structures. For general refinement tasks without structural completion demands (e.g., easy 2D transformations), we focus on the boundary between the object and the background to mitigate unnatural blending. Specifically, we define the boundary mask using morphological dilation~\cite{opencv}:
\[
M_b = \text{Dilation}(M_s) - M_s,
\]
and set $\mathcal{M} = M_b$ to guide the refinement process.

Since our framework's components can be seamlessly replaced by other tools, when undesired content is generated in the source region, we can further refine it in Step~3 by simply including the problematic region in $\mathcal{M}$ during LP.

\section{Evaluation and Baselines}
This section provides additional details on the evaluation metrics and the implementation of baseline methods used in our experiments.

\subsection{Metrics}
Details of how to calculate the metrics: (1) FID~\cite{FID}. The implementation of pytorch-fid~\cite{pytorch-FID} is used, specifically by extracting feature vectors from Inception-v3~\cite{inception} and computing the Fréchet distance (FID) between the distributions of the edited results and 2,000 randomly selected source images. Furthermore, we replace the Inception features with DINO features (DINO) and use kernel distance (KD) instead of the Fréchet distance to mitigate the bias inherent in the original FID, enabling a more comprehensive evaluation of image quality. (2) Subject Consistency (SC). It measures the consistency of the foreground subject between source image $I_s$ and cgenetated image $I_t$. Given the source mask $M_s$ and the target mask $M_t$, SC value is calculated as:
\[
V_{SC} = \cos\left( \mathbf{F}_{\text{DINO}}[I_s \cdot M_s ],  \mathbf{F}_{\text{DINO}}[I_t \cdot M_t ] \right),
\]
where $\cos\left(\cdot\right)$ indicates the cosine distance and $\mathbf{F}_{\text{DINO}}$ represents the DINO~\cite{DINO} feature of a image. (3) Background Consistency (BC). It measures the consistency of the background and is calculated similarly to SC value but with the CLIP~\cite{CLIP} feature $\mathbf{F}_{\text{CLIP}}$:
\[
V_{BC} = \cos\left( \mathbf{F}_{\text{CLIP}}[I_s \cdot M_{bg}],  \mathbf{F}_{\text{CLIP}}[I_t \cdot M_{bg} ] \right),
\]
where $M_{bg} = 1-(M_s\cup M_t)$ represents the mask of background. The settings of using different features are from VBench~\cite{vbench}. (4) Warp Error (WE). We follow the implementation described in GeoDiffuser~\cite{geodiffuser} to get the warped image $I_w$ and compute the L1 error $L_1\left(\cdot\right)$:
\[
V_{WE} = L_1\left(
I_t \cdot M_t, 
I_w \cdot M_t \right).
\]
(5) Mean Distance (MD). We first find interest points $P_i$ on the source image using SiFT~\cite{SiFT}, which are transformed with the transform $\mathcal{F}$ to obtain target positions $P_t$ and used to locate the corresponding points $P_c$ in the edited image via DiFT~\cite{DiFT}. The mean distance between $P_t$ and $P_c$ reflects the accuracy of the edit.

\subsection{DragonDiffusion~\cite{Dragon}}

DragonDiffusion supports object moving and resizing in its original implementation. The source mask $M_s$ in our method is one of the inputs of DragonDiffusion to locate the original content position. Instead of the original way of representing edits with drag points, we directly translate the edit parameters into model inputs. By specifying the target mask $M_t$ in our method as its target dragging position, we apply DragonDiffusion to the rotating task. As for 3D-edits, We implement it by setting the drag points in the content dragging demo. All other hyper parameters and optional inputs remain unchanged by default.

\subsection{MotionGuidance~\cite{MotionGuidance}}

The core input to MotionGuidance is the optical flow that represents the edit task. For 2D-edits, we can simply get the optical flow by geometric calculation with the source mask $M_s$ for the object and the editing parameters. For 3D-edits, We use SV3D~\cite{SV3D} to warp the source image and use RAFT~\cite{RAFT} to estimate the optical flow between the warped image and the original. Once getting the source image and the target optical flow, we can implement MotionGuidance in different editing tasks. Notably, MotionGuidance has 500 denoising steps in the default settings and takes an average of 40 minutes per image. Considering the fairness and the cost of the comparison experiment, we set the same 50 steps as the others and use the editing results for the perceptual study. As for qualitative comparisons, the default settings are used.

\subsection{RegionDrag~\cite{RegionDrag}}
RegionDrag requires the indication of source and target regions to perform editing tasks. This method supports 2D moving, scaling, and rotation; however, the 3D editability of RegionDrag is inherently limited due to the input form and optimization strategy. We use the official implementation released by the authors and pass in the source mask $M_s$ and target mask $M_t$ in GeoBench to indicate the source and target edit regions respectively. We use default values for other hyperparameters and optional inputs.

\subsection{Self-Guidance~\cite{Self-Guidance}}

Self-Guidance does not explicitly take in point-based or region-based input. Instead, this method needs textual input to indicate the subject that requires editing. We discover that the official demo code from the authors does not support real image editing; therefore, we follow the implementation described in GeoDiffuser ~\cite{geodiffuser} and use the code base from ~\cite{self-guide-implement}. We first use DDIM Inversion on real images to obtain the starting noise latent. To indicate the desired 2D and 3D transformations, we use the geometric editing instructions from our dataset. Finally, we use Eq.~9 in Self-Guidance~\cite{Self-Guidance} to optimize both the object shape and appearance.

\subsection{DragDiffusion~\cite{dragdiffusion}}
DragDiffusion relies on the user input drag points for geometric image editing. This method gradually drags the source points to the target location by running an iterative optimization process on the image latent in a specific intermediate time step. We directly leverage the original implementation of DragDiffusion by feeding in the source mask $M_s$ and editing parameters to automatically obtain the control points.

\subsection{GeoDiffuser~\cite{geodiffuser}}
GeoDiffuser achieves both 2D and 3D edits by using shared reference and edit attention layers. The attention sharing mechanism will take in geometric transformations and create corresponding losses for image latent optimization. We directly use the original GeoDiffuser implementation for evaluation. We convert transformation parameters from GeoBench into transformation matrix to serve as the model input. In 2D scenarios, we always use a constant depth map for the whole image following GeoDiffuser, while for 3D, we estimate the image depth using Depth Anything. ~\cite{DepthAnything}

\subsection{Diffusion Handles~\cite{Diffusionhandles}}
Diffusion Handles performs geometric edits by leveraging depth maps and camera transformations. Its workflow flow proceeds as follows: first, it conducts null-text inversion using the depth-to-image Stable Diffusion model, then inpaints the foreground region of the target object with LaMa~\cite{Lama}. The inpainted image is subsequently used to estimate the scene's background depth, which is blended with the transformed foreground object. Finally, the edited image is generated using the transformed depth map combined with the transformed activations of the depth-to-image SD model, as detailed in~\cite{Diffusionhandles}. Our implementation is built on the codebase of GeoDiffuser~\cite{geodiffuser}, which adopts the official implementation provided by the authors of Diffusion Handles~\cite{Diffusionhandles}. For parameter compatibility, we adapt our editing parameters to fit the model's input format and use a constant depth setting for 2D editing.

\subsection{DesignEdit~\cite{DesignEdit}}
DesignEdit treats geometric image editing as two sub-tasks: multi-layered image decomposition and multi-layered image fusion. This method segments out the foreground object and background in the latent space, then applies instruction guided latent fusion that pastes the multi-layered latent representations onto a canvas latent. We use the official implementation~\cite{DesignEdit} released by the authors and additionally implement rotation editing on top of it.

\subsection{Inpainting Methods}
We conduct experiments using MAT~\cite{MAT}, LaMa~\cite{Lama}, BrushNet~\cite{Brush}, and Stable Diffusion Inpainting~\cite{stableinpainting}. As mask dilation is a common technique for object removal, we set a dilation factor of 30 on the source mask $M_s$ for Step~2 across all methods, including ours, to minimize object remnants caused by imperfect masks. Additionally, we use the conditional prompt ``empty scene'' for guidance and fix the random seed in all comparative and ablation experiments to ensure fair and reproducible results.

For the structure completion task, we use manually drawn masks to guide region completion and employ object-specific text prompts from our GeoBench dataset. For general refinement tasks without structure completion, we use boundary masks, as described in the Local Perturbation (LP) section, to guide the refinement process.

\section{Details of GeoBench} \label{sec:more_on_GeoBench}

In the main paper, we introduced \textbf{GeoBench}, a benchmark for evaluating geometric editing methods. This section details its data generation pipeline, including mask generation, label extraction, editing instruction creation, and manual supervision, along with dataset statistics.

\vspace{1mm}\noindent\textbf{Mask Generation.}~Accurate object masks are essential for distinguishing foreground from background and enabling precise region editing. We use the segmentation tool SAM~\cite{SAM} to automatically segment datasets sourced from PIE-Bench~\cite{pie-BENCH} and Subjects200K~\cite{Subjects200k}, which consist of real and synthetic images featuring prominent objects suitable for geometric editing. However, SAM faces challenges in segmentation granularity (e.g., instance, panoptic, part-level) and may not align with our editing requirements. To overcome this, we use SemanticSAM~\cite{semanticSAM}, which allows manual adjustment of segmentation granularity, ensuring more precise and relevant masks. 
Post-processing first applies algorithm-based filtering: tiny masks are discarded, and images with excessive masks (over 50 per image, indicating crowded scenes with limited space for geometric edits) are excluded, where the former is defined as:
\begin{equation*}
\text{\tt tiny mask} \iff \frac{\sum M}{h \times w} < 0.001
\end{equation*}  
where $M$ denotes the object mask, $h$ is the image height, and $w$ is the image width. Remaining masks then undergo manual selection to retain the most prominent (e.g., texture-clear, non-blurry) and editable foreground objects (non-overlapping with other objects, not truncated by image boundaries).

\vspace{1mm}\noindent\textbf{Label Generation.}~We use ShareGPT4V~\cite{SHARE4V} to generate detailed image descriptions and an algorithm based on CLIP similarity~\cite{CLIP, SCLIP} to extract object labels. By comparing the CLIP similarities before and after background inpainting, we select the top $k$ most related labels and send them to human filtering for final verification. Here, we set $k=5$ as the most appropriate label consistently appears in the top 5. An alternative approach, GroundingSAM~\cite{GroundingSAM}, offers simultaneous mask and label generation but shares the same limitations regarding segmentation granularity and error accumulation, often resulting in irrelevant or overly coarse masks.

\vspace{1mm}\noindent\textbf{Editing Instruction Generation.}~To enable diverse and multi-level edits, we design a range of editing prompts and randomly generate instructions for each image, categorized into three difficulty levels: easy, medium, and hard, as detailed in \cref{tab_sup:category}. Instructions include transformations such as moving (eight directions: up, down, left, right, and diagonals), resizing (zooming in/out), and rotating (clockwise, counterclockwise, and 3D rotations).
For 3D rotation instructions, we validate direct depth estimation-based transformations and exclude samples with unreliable depth estimates.
We also exclude instructions that result in objects extending beyond the image boundaries. Additionally, we manually identify cases requiring structural completion and create a dedicated subset for these tasks, along with manually drawn completion masks for each image.

\vspace{1mm}\noindent \textbf{Dataset Statistics.}~{GeoBench} comprises 811 source images and 5,988 editing instructions, including 2,267 easy, 2,075 medium, and 1,646 hard edits. The dataset is further divided into three subsets: (1) general 2D edits (5,677 instructions), (2) 3D edits (190 instructions), and (3) a manually annotated subset for structural completion tasks (121 instructions).

\begin{table}[tb]
\renewcommand{\arraystretch}{1.05}
\scriptsize
\setlength{\tabcolsep}{2mm}
\centering
\caption{Parameter ranges for editing operations across three difficulty levels.}
\label{tab_sup:category}
\begin{tabular}{lccccc}
\toprule
Difficulty & 
\makecell{2D Rotate \\ (\textdegree)} & 
\makecell{Move \\ (Frac. of \\ Img Size)} & 
\makecell{Resize \\ (Enlarge)} & 
\makecell{Resize \\ (Shrink)} & 
\makecell{3D Rotate \\ (\textdegree)} \\ 
\midrule
Easy   & 5--10  & 0.05--0.1 & 1.1--1.3 & 0.8--0.9 & 5--10  \\
Medium & 10--20 & 0.1--0.2  & 1.3--1.5 & 0.6--0.8 & 15--20 \\
Hard   & 20--40 & 0.2--0.4  & 1.5--3.0 & 0.4--0.6 & 25--40 \\
\bottomrule
\end{tabular}
\end{table}
\section{User Study Details}
\begin{figure}[htb]
  \centering
  \subfloat[Home page
    \label{fig_sup:home}]{
    \includegraphics[width=0.4\textwidth]{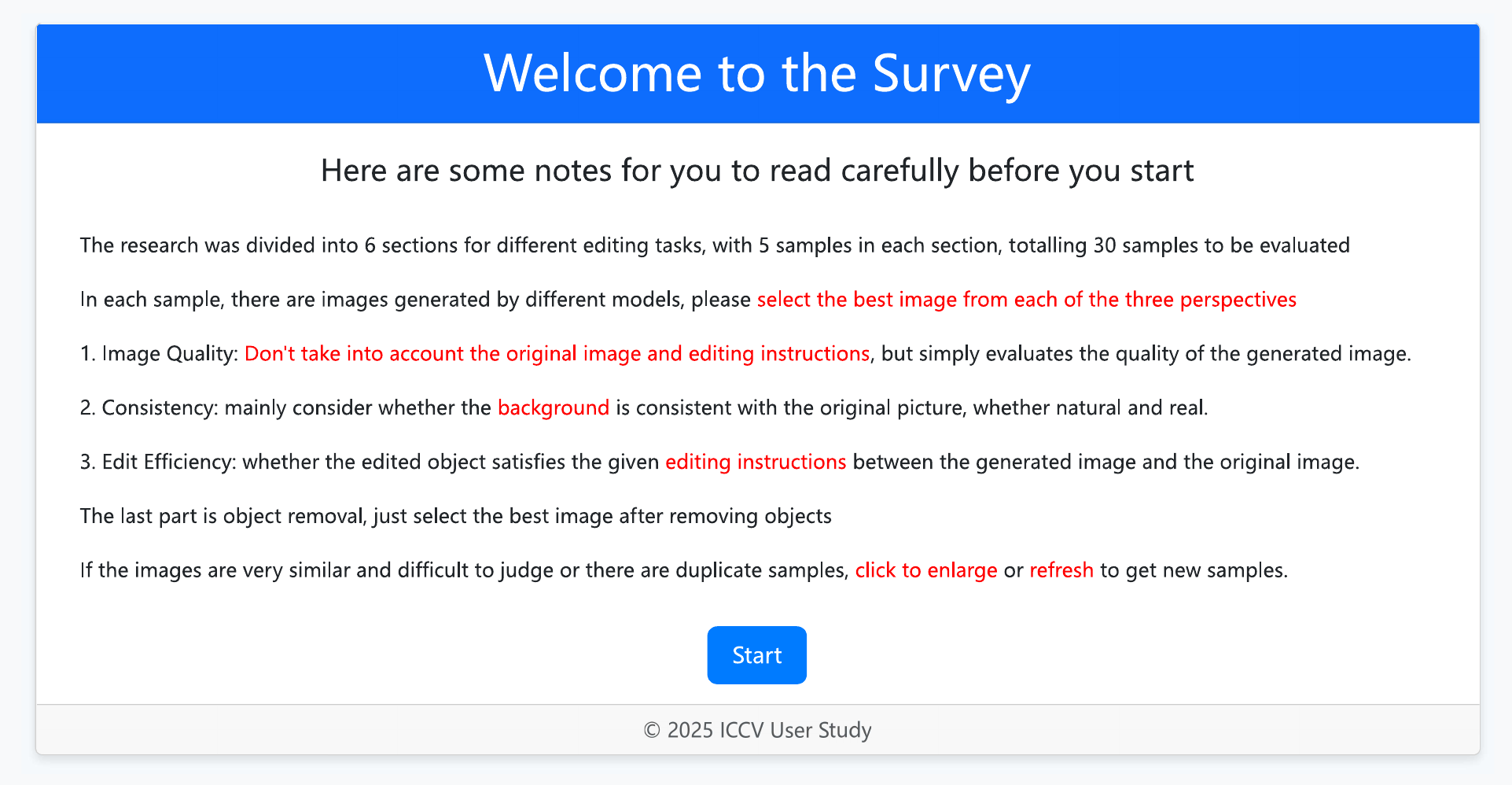}} 
 \vspace{5mm}
  \subfloat[Vote page
    \label{fig_sup:vote}]{
    \includegraphics[width=0.4\textwidth]{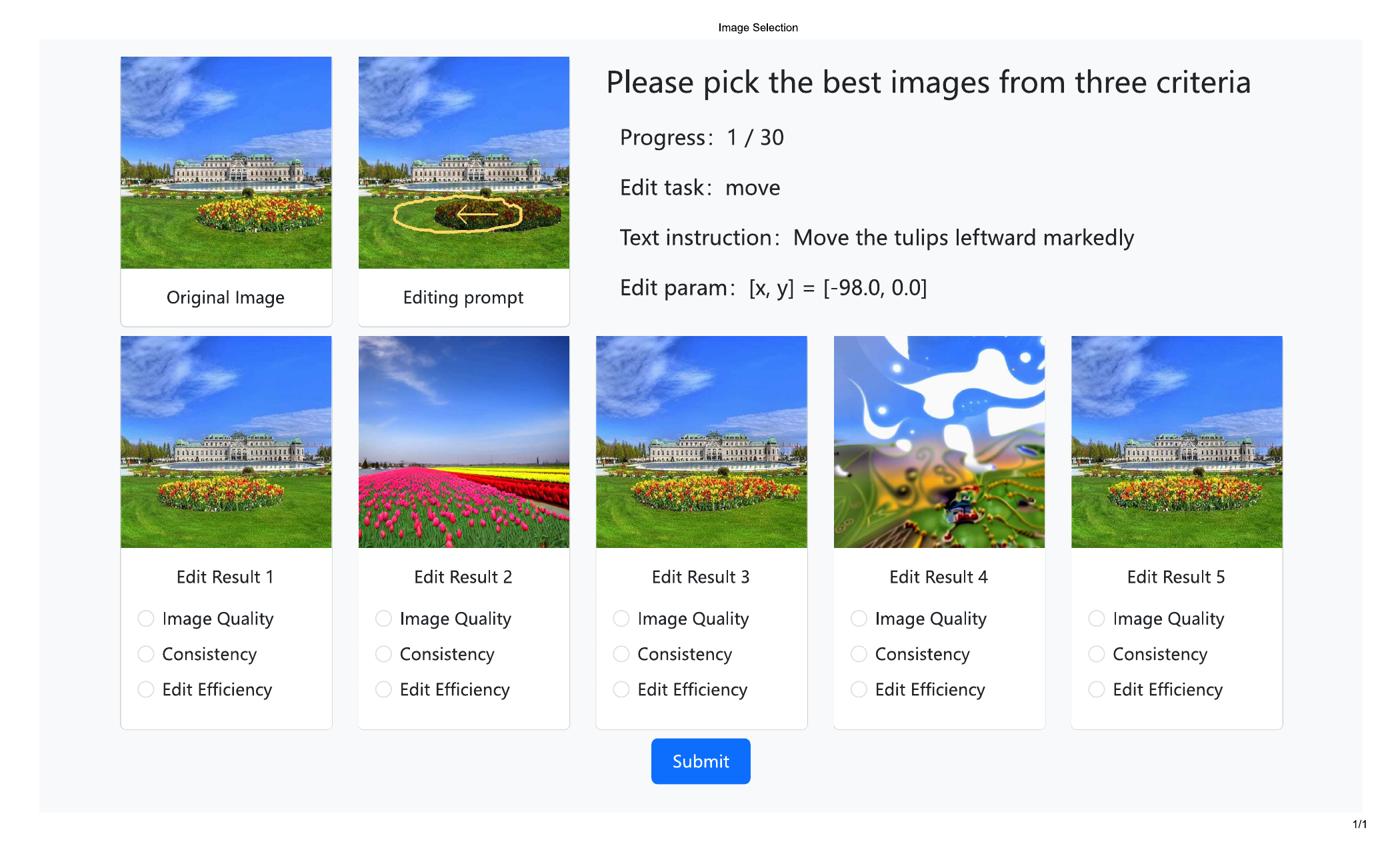}}
     \vspace{5mm} 
  \caption{Screen shots of the website for user study.}
  \label{fig_sup:web}
  \vspace{5mm} 
\end{figure}

A website was built for the user study and we recruited 35 participants with diverse backgrounds in computer vision to vote online. The home page (\cref{fig_sup:web}~\subref{fig_sup:home}) explains the voting task and provides guidelines for participants. The survey includes 6 sections (Move, Rotate, Resize, 3D-Edits, Region Refinement, Region Inpainting), with 5 samples per section, totaling 30 evaluations and the first three sections are all 2D-edits. The Start button redirects users to the vote page (\cref{fig_sup:web}~\subref{fig_sup:vote}), which presents a random editing sample, including the original image, the visualisation of editing prompt, anonymous editing results of our method and comparative models in a random order and the textual information about the sample. Figure  shows visualization results of specific tasks (Move, Rotate, Resize) in 2D-edits.

\begin{figure}[htb]
  \centering
  \includegraphics[width=0.47\textwidth]{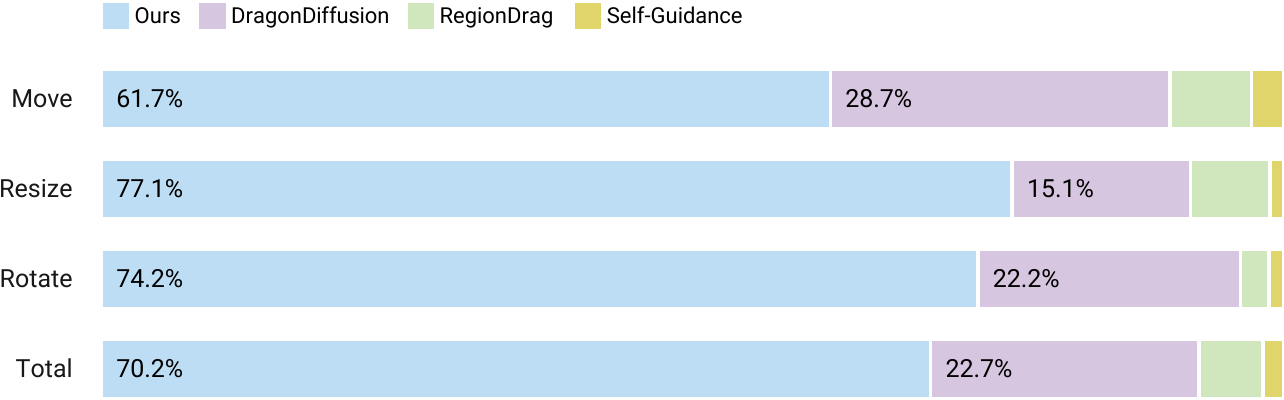}
  \caption{Visualization results of perceptual study in 2D-edits (Move, Rotate, Resize).}
  \label{fig_sup:2d_edit}
\end{figure}

We compared (1) editing models (DragonDiffusion~\cite{Dragon}, RegionDrag~\cite{RegionDrag}, Self-Guidance~\cite{Self-Guidance}, MotionGuidance~\cite{MotionGuidance}) in 2D-edits and 3D-edits, (2) inpainting models (BrushNet~\cite{Brush}, SD-inpainting-v1.5~\cite{SD}, LaMa~\cite{Lama}, MAT~\cite{MAT}) in the region inpainting task, (3) BrushNet~\cite{Brush}, SD-inpainting-v1.5~\cite{SD} and editing models in the Region Refinement task. 35 participants picked the best image from three criteria and submitted it, generating 2,622 valid votes (\cref{tab_sup:vote}). In the final section, participants selected one best image (instead of three criteria) in the region inpainting task, with the voting number as one-third of the other sections.~\cref{fig_sup:2d_edit} shows voting statistics in the 2D-edits and 3-edits from different criteria, only the editing models are counted.

\begin{figure*}[htb]
  \centering
  \includegraphics[width=0.95\textwidth]{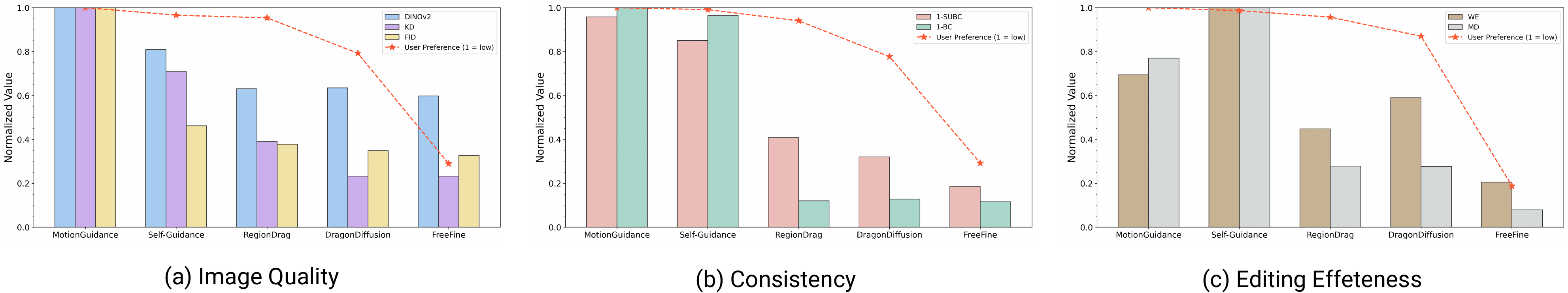}
  \caption{Assessment of the Alignment between Metrics from the Main Paper and User Preferences across Three Dimensions.}
  \label{fig_sup:metric_val}
\end{figure*}

We also validate the alignment between the metrics used in the main paper and user preferences across three key dimensions: Image Quality, Consistency, and Editing Effectiveness, as seen in the~\cref{fig_sup:metric_val} with data from \cref{tab_sup:criteria}. 
The seven metrics generally correlate strongly with human preference. However, the WE metric under Editing Effectiveness shows some degree of misalignment, as evidenced by the human preference ratings for RegionDrag and DragonDiffusion. Developing more robust quantitative metrics to better measure editing effectiveness remains an important direction for future work.

\begin{table}[tb]
\footnotesize
\setlength{\tabcolsep}{1.2mm}
\centering
\caption{Voting statistics in the 2D-edits and 3-edits from different criteria, only the editing models are counted.}
\label{tab_sup:criteria}
\begin{tabular}{lcccc}
\toprule
    Method & 
    \makecell{Image \\ Quality} & Consistency & \makecell{Editing \\ Effectiveness} & Total \\ 
    \midrule
    Ours & \textbf{475} & \textbf{473} & \textbf{543} & \textbf{1491} \\
    DragonDiffusion \cite{Dragon} & 139 & 149 & 87 & 375 \\
    RegionDrag \cite{RegionDrag} & 31 & 40 & 29 & 100 \\
    Self-Guidance \cite{Self-Guidance} & 23 & 6 & 9 & 38 \\
    MotionGuidance \cite{MotionGuidance} & 0 & 0 & 0 & 0 \\
    \midrule
    Total & 668 & 668 & 668 & 2004 \\
    \bottomrule
\end{tabular}
\end{table}

\begin{table}[hbt]
    \tiny
    \setlength{\tabcolsep}{1.2mm}  
    \centering
    \caption{Voting statistics in different editing tasks. The blank cells indicate that the model was not compared in the task.} 
    \label{tab_sup:vote}
    \begin{tabular}{lcccccccc}
    \toprule
        \multirow{2}{*}{Method} & 
        \multicolumn{4}{c}{2D-Edits} & 
        \multirow{2}{*}{3D-Edits} & 
        \multirow{2}{*}{ \makecell{Region \\ Refinement}} & 
        \multirow{2}{*}{\makecell{Region \\ Inpainting}} & 
        \multirow{2}{*}{Total} \\ 
        \cmidrule(lr){2-5}
        ~ & Move & Resize & Rotate & Total & ~ & ~ & ~ & ~ \\ \midrule
        
        Ours & \textbf{374} & \textbf{347} & \textbf{365} & \textbf{1086} & \textbf{405} & \textbf{258} & 37 & 1786 \\ 

        DragonDiffusion\cite{Dragon} & 174 & 68 & 109 & 351 & 24 & 50 & ~ & 425 \\
        RegionDrag\cite{RegionDrag} & 42 & 30 & 12 & 84 & 16 & 12 & ~ & 112 \\
        Self-Guidance\cite{Self-Guidance} & 16 & 5 & 6 & 27 & 11 & 3 & ~ & 41 \\ 
        MotionGuidance\cite{MotionGuidance} & 0 & 0 & 0 & 0 & 0 & 0 & ~ & 0 \\

        \midrule
        
        BrushNet\cite{Brush} & ~ & ~ & ~ & ~ & ~ & 89 & 22 & 111 \\
        SD-inpaint-v1.5\cite{SD} & ~ & ~ & ~ & ~ & ~ & 56 & 21 & 77 \\ 
        LaMa\cite{Lama} & ~ & ~ & ~ & ~ & ~ & ~ & \textbf{52} & 52 \\ 
        MAT\cite{MAT} & ~ & ~ & ~ & ~ & ~ & ~ & 18 & 18 \\
        
        \midrule
        Total & 606 & 450 & 492 & 1548 & 456 & 468 & 150 & 2622 \\ \bottomrule
    \end{tabular}
\end{table}
\begin{figure*}[htb]
  \centering
  \includegraphics[width=0.95\linewidth]{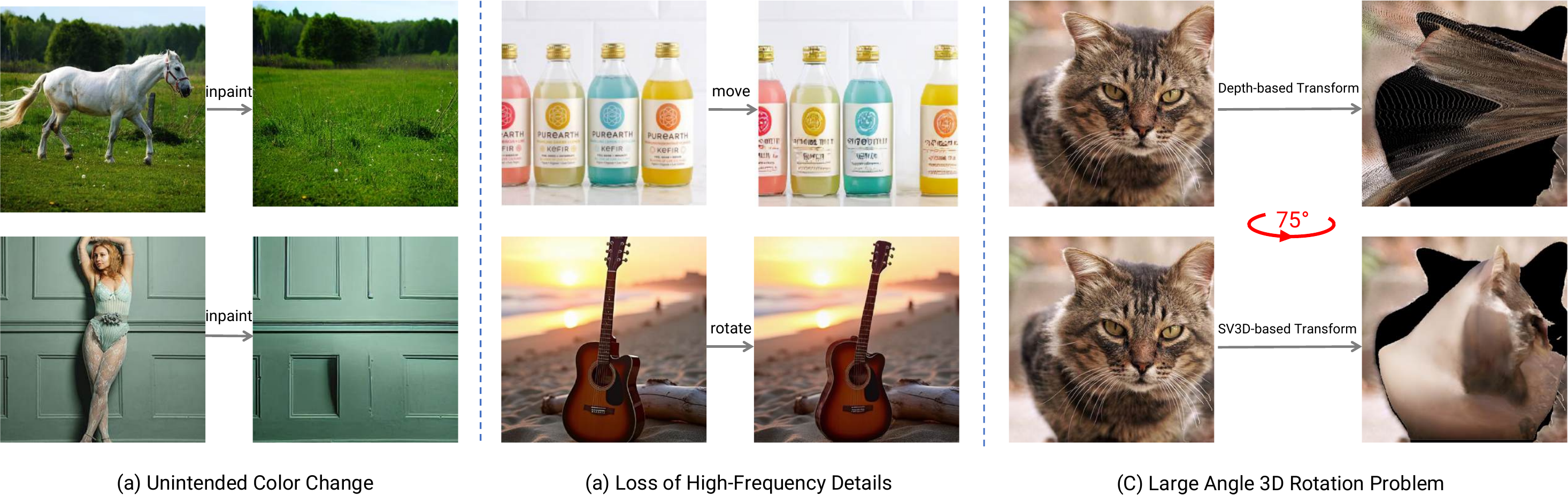}
  \caption{Visualization of failure cases.}
  \label{fig_sup:fail_cases}
\end{figure*}

\section{Failure Cases and Limitations}

While our method achieves strong performance across a variety of geometric editing tasks, it still faces certain failure cases and limitations, which are detailed below.

\subsection{Failure Cases}
Failure cases are illustrated in Fig.~\ref{fig_sup:fail_cases}:  
(1) Unintended color changes may occasionally occur in the background during editing. This likely stems from the need for early denoising steps in Step~2 (implemented to avoid interference from the source region’s context), which can cause information loss and hinder color consistency.  
(2) Fine details such as tiny text or guitar strings may appear blurry, as preserving high-frequency features remains a technical challenge.  
(3) Despite supporting larger 3D rotations via SV3D, our method still struggles with large-angle 3D rotations for many objects. This is due to suboptimal coarse edits from lifting models, creating significant barriers to effective refinement. We hope, however, that FreeFine’s inherent flexibility will help mitigate these issues through the integration of stronger foundation models in future work.  
Addressing these failure cases is a key direction for our future work.

\subsection{Limitations}
While our method shows strong performance, it still has several limitations.
First, it relies on user input in two aspects: (1) it requires user-provided masks for editing guidance, and (2) structure completion tasks depend on manual intervention to guide the process. Developing automated pipelines for these steps would reduce manual effort.  
Second, for complex 3D edits, our method depends on depth estimates or other 3D models. Removing this requirement, if possible, would make the system more broadly applicable.  
Third, similar to most diffusion-based methods, our approach has relatively high computational costs compared to feed-forward models like GANs. Adopting more efficient sampling strategies could lower inference costs.


\end{document}